\documentclass{article}

\usepackage{arxiv}

\usepackage[utf8]{inputenc} % allow utf-8 input
\usepackage[T1]{fontenc}    % use 8-bit T1 fonts
\usepackage{hyperref}       % hyperlinks
\usepackage{url}            % simple URL typesetting
\usepackage{booktabs}       % professional-quality tables
\usepackage{pifont}% http://ctan.org/pkg/pifont
\usepackage[section]{placeins}
\usepackage{makecell}

\usepackage{rotating}

\usepackage{amsfonts}       % blackboard math symbols
\usepackage{nicefrac}       % compact symbols for 1/2, etc.
\usepackage{microtype}      % microtypography
\usepackage{lipsum}		% Can be removed after putting your text content
\usepackage{graphicx}
\usepackage{natbib}
\usepackage{doi}
\usepackage{url}

\usepackage{breakurl}

\newcommand{\cmark}{\ding{51}}%

\title{Visualizing and Explaining Language Models}

\author{\href{https://orcid.org/0000-0002-3439-4736}{\includegraphics[scale=0.06]{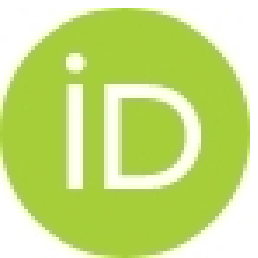}\hspace*{1mm}
Adrian M.P. Bra\c{s}oveanu}\\
	Modul University Vienna\\
	and Modul Technology GmbH\\
	Am Kahlenberg 1, 1190, Vienna, AT \\
	\texttt{adrian.brasoveanu@modul.ac.at} \\
	\And
	\href{https://orcid.org/0000-0002-6015-3151}{\includegraphics[scale=0.06]{orcid.eps}\hspace*{1mm}R\u{a}zvan Andonie} \\
	Central Washington University\\
	400 E University Way,\\
	Ellensburg, WA 98926, USA, \\ 
	and Transilvania University\\
    Bulevardul Eroilor 29, Brașov 500036, Romania \\ 
	\texttt{razvan.andonie@cwu.edu} \\
}

\renewcommand{\shorttitle}{Visualizing and Explaining Language Models}

\hypersetup{
pdftitle={A template for the arxiv style},
pdfsubject={q-bio.NC, q-bio.QM},
pdfauthor={David S.~Hippocampus, Elias D.~Striatum},
pdfkeywords={First keyword, Second keyword, More},
}

\begin{document}
\maketitle

\begin{abstract}
During the last decade, Natural Language Processing has become, after Computer Vision, the second field of Artificial Intelligence that was massively changed by the advent of Deep Learning. Regardless of the architecture, the language models of the day need to be able to process or generate text, as well as predict missing words, sentences or relations depending on the task. Due to their black-box nature, such models are difficult to interpret and explain to third parties. Visualization is often the bridge that language model designers use to explain their work, as the coloring of the salient words and phrases, clustering or neuron activations can be used to quickly understand the underlying models. This paper showcases the techniques used in some of the most popular Deep Learning for NLP visualizations, with a special focus on interpretability and explainability.	
\end{abstract}

% keywords can be removed

\section{Introduction}
\label{sec:introduction}

Deep Learning (DL) models applied on texts need to cover the morphological, syntactic, semantic and pragmatic layers. Crafting networks that operate on so many levels is a challenging task due to the sparseness of the training data. Such networks have been traditionally called Language Models (LMs) \cite{DBLP:conf/sigir/PonteC98}. The early iterations were based on statistical models \cite{DBLP:conf/sigir/PonteC98}, whereas the latest iterations use neural networks and embeddings. Current LMs are trained on large corpora and are generally sparse. Most current popular LMs are based on Transformer architectures \cite{DBLP:conf/nips/BrownMRSKDNSSAA20}.

The first implementation of a Transformer network \cite{DBLP:conf/nips/VaswaniSPUJGKP17} proved that it was possible to design networks that achieve good results for Natural Language Processing (NLP) tasks with a set of multiple sequential attention layers. A Transformer contains a series of self-attention layers that are distributed through its various components. Self-attention is an attention mechanism that computes a representation of a sequence from a set of different positions of the same sequence. The Transformer model itself is simple and consists from pairs of encoders and decoders. Encoders encapsulate layers of self-attention coupled with feed-forward layers, whereas decoders encapsulate self-attention layers followed by encoder-decoder attention and feed-forward layers. The attention computation is done in parallel and the results are then combined. The result is termed a multi-head attention, and it provides the model with the ability to orchestrate information from different representation subspaces (e.g., multiple weight matrices) at various positions (e.g., different words in a sentence) \cite{DBLP:conf/nips/VaswaniSPUJGKP17}. Its outputs are fed either to other encoders or into decoders, depending on the architecture. There is no fixed number of encoders and decoders which can be included in this architecture, but they will typically be paired (e.g., 10 encoders and 10 decoders). In newer architectures, encoders and decoders can also be used for different tasks (e.g., encoder for Question Answering, and decoder for Text Comprehension) \cite{DBLP:journals/corr/abs-2003-11755}. While the model was initially developed for machine translation tasks, it has been tested on multiple domains and was demonstrated to work well. 

During the last three years, hundreds of papers and LMs inspired by Transformers were published, the best-known being BERT \cite{DBLP:conf/naacl/DevlinCLT19}, RoBERTa \cite{DBLP:journals/corr/abs-1907-11692}, AlBERT \cite{DBLP:conf/iclr/LanCGGSS20}, XLNet \cite{DBLP:conf/nips/YangDYCSL19}, DistilBERT \cite{DBLP:journals/corr/abs-1910-01108}, and Reformer \cite{DBLP:conf/iclr/KitaevKL20}. Some of the most popular Transformer models are included in the Transformers library, maintained by HuggingFace \cite{DBLP:journals/corr/abs-1910-03771}. 

Many of these models are complex and include significant architectural improvements compared to the early Transformer and BERT models. Explaining their information processing flow and results is therefore difficult, and a convenient and very actual approach is visualization. Our survey is focused on visualization techniques used to explain LMs. We investigate two large tool classes: (i) model-agnostic tools that can be used to explain BERT predictions; and (ii) custom visualizations that are focused only on explaining the inner workings of LMs based on neural networks. An early version of this survey was published a year ago, but it was focused only on visualizing Transformer networks \cite{BrasoveanuA20}. We have since extended the material to include new articles about Transformer visualizations, other types of networks, as well as an extended section about model-agnostics AI libraries that are focused on interpretability and explainability for NLP.

In this survey, we look at the visualization of several types of LMs based on DL networks, review the basic charts and patterns present in them and try to understand the basic methodology that was used to produce these visual representations. The rest of the paper is organized as follows: Section~\ref{sec:background} presents the motivation and methodology of this survey.  Section~\ref{sec:visualization} showcases the two classes of tools, whereas Section~\ref{sec:discussion} discusses the various findings. The paper concludes with some thoughts on the future of this class of visualizations.

\section{Background and Methodology}
\label{sec:background}

The need to quickly update NLP models in case of unforeseen events suggests that developers will be well-served by explainable AI and visualization libraries, especially since debugging Transformers is a complex task. Visualizations are particularly important, as they help us debug the various problems that such models exhibit and which can only be discovered through large-scale analyses.

Traditional visualization libraries are based on the classic grammar of graphics philosophy \cite{DBLP:books/daglib/0024564} which is focused on the idea that visualizations are compositional by design. They provide various \textit{visualization primitives} like circles or squares and \textit{a set of operations} that can be applied on top of these primitives to create more complex shapes or animations. Unfortunately, such traditional visualization libraries like D3.js \cite{DBLP:journals/tvcg/BostockOH11}, Vega \cite{DBLP:journals/tvcg/SatyanarayanMWH17} or Tableau\footnote{www.tableau.com}, do not
offer specific functions for visualizing feature spaces, neural network layers or support for iterative design
space exploration \cite{DBLP:journals/tvcg/ParkKLCDE18} when designing AI models. What this means is that for
AI tasks, a lot of the functionality will have to be developed from scratch. 

When visualizing more complex models like those built with Transformers, we typically need to understand all the facets of the problem, from the data and training procedure, to the input, network layers, weights or the outputs of the neural network. The outputs are the core of explainability, as people will not use the networks in commercial products if they can't explain how the outputs were obtained in the first place. What is also beneficial is to highlight the paths that lead to certain outputs, as this illuminates the features or parts of the models that may need to be changed to achieve the desired results. This can sometimes be accomplished by using model-agnostic tools specifically built for benchmarking or hyperparameter scoring, such as \textit{Weights and Biases}. We include such tools in our survey if examples of how to use them for visualizing Transformers already exist, either in scientific papers or other types of media posts (Medium posts, GitHub, etc.). 

The second big class of visualizations discussed in this article is, naturally, the class of visualizations specifically built around Transformers, either for explaining it (like ExBERT \cite{DBLP:conf/acl/HooverSG20}), or for explaining certain model specific attributes (like embeddings or attention maps \cite{DBLP:conf/acl/Vig19}).

We selected the libraries and visualizations presented here by reviewing the standard Computer Science (CS) publication libraries (e.g., IEEE, ACM, Elsevier, Springer, Wiley), but also online media posts (YouTube, Medium, GitHub and arXiv). In this extremely dynamic research field, some articles might be published on arXiv even up to a year before they are accepted for publication in a traditional conference or journal, time in which they might already garner hundreds of citations. The original BERT article \cite{DBLP:conf/naacl/DevlinCLT19} and also one of the first articles that used visualization to explain it \cite{DBLP:journals/corr/abs-1906-04341} were cited over a hundred times before being published in conference proceedings.\footnote{Article \cite{DBLP:journals/corr/abs-1906-04341} has garnered 149 citations at the moment of the submission, before being published in a conference or journal.} 

When testing new models, benchmarking and fine-tuning are the two operations where we might spend the most time, as even if the scores are good, we might want to try different hyperparameter settings (e.g., learning rate, number of epochs, batch size, etc) \cite{DBLP:journals/ijccc/FloreaA19}. A \textit{hyperparameter sweep} (or trial) is a central notion in both hyperparameter optimization and benchmarking. It involves running one or multiple models with different values for their hyperparameters. Since quite often the main goal behind running such sweeps is improving existing models, but it is not necessarily related to interpretability and explainability, we decided to include a minimal number of such libraries here. 

\textit{TensorBoard}\footnote{https://www.tensorflow.org/tensorboard} is a specialized dashboard deployed with Google TensorFlow that covers the basic visualization needs for ML experiments, from tracking, computing and visualizing metrics, to model profiling and embeddings. It is not necessarily a good tool for creating custom visualization or for explaining results, but it can be a good tool for improving accuracy. It is sometimes also used with TensorFlow's competing libraries like PyTorch or FastAI. \textit{Neptune}\footnote{https://neptune.ai/} is an open-source ML benchmarking suite deployed for a variety of collaborative benchmarking tasks, including notebook tracking. \textit{Sacred}\footnote{https://github.com/IDSIA/sacred} and \textit{Comet.ml}\footnote{https://www.comet.ml/site/} are Neptune alternatives that provide basic charting capabilities and dedicated dashboards. \textit{Weights and Biases}\footnote{https://www.wandb.com/} provides perhaps the largest sets of visualization and customization capabilities. It comes packed with advanced visualizations that include parallel coordinates \cite{DBLP:journals/cse/HeinrichW15}, perhaps the best method to navigate hyperparameter sweeps. It is the easiest and the most agile solution to integrate with production code or Jupyter notebooks out of all the ones mentioned here. \textit{Ray} \cite{DBLP:conf/osdi/MoritzNWTLLEYPJ18}, a distributed benchmarking framework that contains its own fine-tuning engine called \textit{Tune} \cite{DBLP:journals/corr/abs-1807-05118} is popular for optimizing Transformers.

Due to space limitations, we resume ourselves to discussing only the most interesting visualizations, especially in the model-agnostic visualization section, as otherwise this article could easily become an entire book.

\section{Visualizing Language Models}
\label{sec:visualization}

Language models are difficult to train for a multitude of reasons, including (but not limited to) cost, time or carbon footprint \cite{DBLP:conf/acl/StrubellGM19}. Most of the LMs need to be trained on GPUs or TPU pods for days or weeks. Due to their generalization capabilities, such LMs can reliably estimate the actions for which they have a reasonable number of examples in their training datasets, whereas in cases with fewer examples they might overestimate the predicted actions therefore inserting some bias \cite{DBLP:conf/coling/ShwartzC20}. Debugging or retraining such models therefore becomes a necessity, even if the costs of such operations are still high. Visualization is just one of the methods that can help us explain such large LMs, especially since it is often combined with linguistics or statistics. Explaining the results in plain English should be what we are aiming for when we build new LMs, but this may sometimes require additional steps. An interpretation of the results, for example, would generally depend on the target domain (e.g., medicine, law, etc) \cite{DBLP:journals/nca/Vellido20}, as in some cases a complex reasoning process (e.g., compliance with local or international regulations) may need to be applied before selecting the right words for an explanation. The visualization will essentially highlight the intermediary steps (e.g., the components that lead to a side effect for medication or a legal aspect in a certain jurisdiction) required to create a basic interpretation, and therefore it is often kept minimal and visualized features or processes are carefully selected. If it is easy to navigate the various information pathways and understand the results and their interpretations (e.g., where they may lead us), it may be safe to call the respective visualizations explainable.

Using visualization to explain the AI processes is an expanding research field. The main idea behind AI user interfaces should be to augment and expand user capabilities, rather than replace intelligence \cite{DBLP:journals/pnas/Heer19}. While not necessarily needed to understand the next section, several recent surveys about visualizations and DL can help provide additional context to the interested readers. We particularly recommend the following: the introduction on how Convolution Neural Networks "see" the world from \cite{DBLP:journals/mfc/QinYLC18}, the discussion on visual interpretability from \cite{DBLP:journals/jzusc/ZhangZ18}, and the discussion on the importance of visualizing features from \cite{DBLP:series/lncs/NguyenYC19}. 

\subsection{Model-Agnostic Explainable AI Tools}
\label{sec:model-agnostic-xai}

Explainable\footnote{\textit{Explainable} points to the idea of describing or explaining in an intuitive manner, via charts or tables, the prediction of an algorithm.}  AI (XAI) is the key to enterprise adoption of the current wave of AI technologies, from vision to NLP and symbolic computation. An early XAI survey \cite{DBLP:journals/tvcg/HohmanKPC19} describes methods through which visualizations can be turned into explanations for the AI models and goes on to define the terminology of the field. Early XAI libraries focused on visualizing ML features, whereas recent libraries are focused on visualizing embeddings, attention maps or various neural network layers \cite{DBLP:series/lncs/11700}.

Traditionally, the first step towards transparency was to describe the contribution of each feature to the final result \cite{DBLP:journals/jmlr/GuyonE03}. This often lead to partial explanations of the results, as in reality if the models themselves were black-box, knowing the name of the features was not in itself enough.  Output is definitely the most important part that we would like to explain, but not the only one. To create full explanations, we need to be able to explain the entire process from its input, to its various transformations (e.g., layers), training process and output. Explanations also need to be able to reflect state changes. For example, when computing Shapley values feature contributions are combined, and then a score that signifies the feature importance within that set of features is generated. If features are added or removed from this bundle, the Shapley value for a particular feature will change accordingly. This dynamic nature of the explainability is rarely explained, but it is one of the reason why visualizations in particular are a good fit for creating explanations in the first place.

Some early model-agnostic XAI libraries that were applied to NLP and Transformers visualizations include LIME \cite{DBLP:conf/kdd/Ribeiro0G16} and SHapley Additive exPlanations (SHAP) \cite{DBLP:conf/nips/LundbergL17}. The later was introduced to unify multiple explanation methods into a single model for interpreting predictions. Both SHAP and LIME can be used with classical ML libraries like scikit-learn and XGBoost, as well as with modern DL libraries like PyTorch or Keras / TensorFlow. SHAP provides visualizations for summary and dependency plots. Another XAI alternative to SHAP and LIME, ELI5 \cite{DBLP:conf/acl/FanJPGWA19}, is currently routinely used for explaining BERT predictions, and was found to be more secure in case of adversarial network attacks \cite{DBLP:conf/aies/SlackHJSL20}.

The visualizations created with LIME and SHAP are typically restrained to classic charts (e.g., line, bars or word clouds. The summary plots or interaction charts \cite{DBLP:conf/nips/LundbergL17} from SHAP are relatively easy to understand, whereas the more complex force plot charts like feature impact \cite{lundberg2018explainable} are not necessarily easy to use as they require a certain learning time. While the feature impact chart simply plots the expected feature impact with red (features with positive contribution to the prediction result) or blue (features with a likely negative contribution to the result) colors and should in theory be an easy-to-understand chart, there are no direct (e.g., in chart via a legend) explanations on how to interpret the start or end values, or what do the indicators placed on top of various components mean in some cases. The interpretation of such force plot charts is generally missing and people need to read additional documentation to understand the results. This is far from ideal, as, in our opinion, visualizations need to be self-explanatory. 

It can be argued that explainability libraries like SHAP or LIME tend to focus on highlighting correlations or statistical effects rather than features, and are, therefore, less reliable than interpretable models which only showcase a list of features or algorithms that contributed to the results. We consider auditing to encompass both sides of the problem, as  
interpretability and explainability are the key towards understanding and clearing such LMs for deployment in real-world products. Keeping this in mind, we think that focusing on neural network visualization for NLP can only help in this process, as visualizations can help process, select and highlight the most important features included in such models.

Both SHAP and LIME were proven to be easily fooled with adversarial attacks \cite{DBLP:conf/aies/SlackHJSL20}. The idea of deploying biased classifiers for tasks like credit rating, recommendation or search ranking sounds a bit counterintuitive because a single classifier should not be able to do much harm. However, given the fact that such large models typically end up being ensembles, one single classifier can actually lead to severe damage including wrong predictions, different sets of biases and ultimately even different outputs than the ones typically expected from the respective LMs. Not being able to correctly audit such models (e.g., investigate their output and the features that have contributed the most to it) can lead to problems with clients and regulatory agencies. 

The list of attacks that can be perpetrated using LMs is extended every month, and therefore models may need to be periodically tested to assess their suitability for certain tasks. Some of the most robust attacks include: creating token sequences that act like universal adversarial triggers on specific target predictions when concatenated to any input from a dataset \cite{DBLP:conf/emnlp/WallaceFKGS19}; training data extraction attacks in which text sequences like public information or code are extracted from the training corpora and used to attack the trained language model \cite{DBLP:journals/corr/abs-2012-07805}; spelling attacks in which random spellings for well-known words are generated through modifying the gradients during training \cite{DBLP:journals/corr/abs-2003-04985}; hotflip attacks in which a gradient-based embeddings swap is performed to change classification results \cite{DBLP:conf/acl/EbrahimiRLD18} or even textfooler attacks \cite{DBLP:conf/aaai/JinJZS20} in which multiple attributes like embeddings, part-of-speech matches or cosine similarities are used to perform a counter-fitted embeddings swap to fool untargeted classifiers and entailment relations. Possibilities to reuse part of the code of these attacks to create new attacks also exists if frameworks like TextAttack \cite{DBLP:conf/emnlp/MorrisLYGJQ20} or OpenAttack \cite{DBLP:journals/corr/abs-2009-09191} are used. Sometimes these kinds of adversarial attacks can also be used to improve results, as demonstrated by improving aspect-based sentiment analysis tasks by creating artificial sentences \cite{DBLP:conf/icpr/KarimiR020} or by using various BERT attribution patterns (e.g., pruned self-attention heads from a certain task) as adversaries \cite{DBLP:journals/corr/abs-2004-11207}. To understand these attacks, visualization can be a useful tool. For example, Chen \cite{DBLP:journals/corr/abs-2006-01043} showcases three types of attacks perpetrated at character, word and sentence level using visualizations built around various metrics like accuracy and successful attack ratios. The Interpret framework \cite{DBLP:conf/emnlp/WallaceTWSGS19} uses saliency maps color highlights to showcase defenses against hotflip and untargeted classification attacks. A later publication then shows how almost all interpretations built with Interpret can be manipulated through gradient attacks (e.g., using some large gradients for irrelevant words) \cite{DBLP:conf/emnlp/WangTW020}. However, most of the visualization efforts have been focused on explaining the various neural network activations from Transformer networks, rather than on the various attacks that can fool Transformers, therefore the visualization of such attacks is a relatively nascent area.

An alternative approach to such explainability libraries that can easily pick up wrong signals could be to simply use models designed specifically to be interpretable. The Neural Additive Models (NAMs) combines the features of classic DNNs with the interpretability approach advocated by Generalized Additive Models (GAM) \cite{DBLP:journals/corr/abs-2004-13912}. However, since such models are quite recent, their applicability to NLP has not yet been fully explored. 

Interpretability and explainability are often used with interchangeable meaning. It has to be noted that quite often, interpretability has a domain-specific component \cite{DBLP:journals/nca/Vellido20}, whereas explainability is a more general term. Explainability is the term preferred by Information Visualization designers and researchers, whereas interpretability is generally the term that is preferred by ML researchers, statisticians and mathematicians.

Many other explainable AI libraries use Shapley values for computing feature importance. However, in many cases we were only barely able to discover mentions of their usage for NLP (e.g., DeepExplain\footnote{https://github.com/marcoancona/DeepExplain}), and therefore we decided not to include them in this survey.

\subsection{Visualizing Recurrent Neural Networks for NLP}
\label{sec:rnns-nlp}

When visualizing LMs, it is best to start with the language resources used for their creation, from corpora to embeddings. To uncover biases in such large models we need to study gender differences, disciplines, languages, cultural context or regional and diachronic variations. A good method to include such information and compare resources for language variation is showcased in Fanhauser's work \cite{DBLP:conf/lrec/FankhauserKT14}. It uses a grids, heatmaps and word clouds to provide quick access to large amounts of data about English dialects. Another work uses scatter plots to visualize how and why large corpora differ \cite{DBLP:conf/acl/Kessler17}. Similar methods have later been used for visualizing large-scale embeddings.

Karpathy \cite{DBLP:journals/corr/KarpathyJL15}  proposed a method through which to visually interpret the results of Recurrent Neural Networks (RNNs), Long Short-Term Meories (LSTMs) and Gated Recurrent Units (GRUs). He suggests that using interpretable activations can help navigate longer texts, whereas saturation plots would help showcase the gated units statistics.

\begin{table*}[]
\caption{Articles focused on explaining RNN LMs through visualizations.}
\label{tab:explain-rnns}
\centering
\begin{tabular}{@{}lll@{}} \toprule \\
        Topic & Visualization Subject & Chart Type  \\ \midrule

a) Special topics & & \\ \hline
\hline

\makecell[l]{predicting \\ next word \cite{DBLP:conf/acl/LuoJBG19}} & \makecell[l]{syntactic heights \\ head attentions} & \makecell[l]{parallel lines \\ heatmaps}\\ \hline

\makecell[l]{emergence of units \\ \cite{DBLP:conf/naacl/LakretzKDHDB19}} & \makecell[l]{activations of cells and gates \\ connectivity } & \makecell[l]{line charts \\ connectivity charts} \\ \hline

b) Hidden states & & \\ \hline
\hline

\makecell[l]{visualizations of \\ representations \cite{DBLP:conf/naacl/LiCHJ16}} & \makecell[l]{modification and negation \\ clause composition \\ first-derivative saliency} & \makecell[l]{t-SNE \\ heatmap \\ saliency bars \\ saliency grids} \\ \hline

\makecell[l]{hidden states\\  semantics \cite{DBLP:conf/visualization/SawatzkyBP19} } & \makecell[l]{Predictive Semantic Encodings \\ performance metrics} & \makecell[l]{PSE charts \\ Bar charts}\\ \hline

\makecell[l]{activation of \\ RNNs \cite{DBLP:journals/corr/KarpathyJL15}} 
 & \makecell[l]{cells with interpretable \\ activation} & \makecell[l]{saturation plots \\ bar charts \\ overlap charts}\\ \hline

\makecell[l]{hidden states \\ LSTMVis \cite{DBLP:journals/tvcg/StrobeltGPR18}} & \makecell[l]{phrase \\ selections \\ hidden state patterns} & \makecell[l]{hidden pathways \\ tables with cell activation \\ PCA}\\ \hline

\makecell[l]{hidden states \\ RNNVis \cite{DBLP:conf/ieeevast/MingCZLCSQ17}} 
& \makecell[l]{navigation \\ sentence \\ hidden state \\ word } & \makecell[l]{control panel \\ glyph-based chart \\ state clusters \\ word clusters} \\ \hline

\makecell[l]{hidden states \\ ActiVis \cite{DBLP:journals/tvcg/KahngAKC18}} & \makecell[l]{navigation \\ neuron activation \\ instance selection } & \makecell[l]{model overview \\ heatmaps \\ }
\\ \hline

c) Graph Convolution & & \\ \hline
\hline
%\cite{DBLP:conf/sacmat/FengY20} & 
%phishing detection \\\hline
\makecell[l]{inductive text  \\ classification \cite{DBLP:conf/acl/ZhangYCWWW20}} & \makecell[l]{attention \\ performance metrics} & \makecell[l]{attention map \\ line charts} \\ \hline

\makecell[l]{unsupervised domain \\ adaptation \cite{DBLP:conf/www/WuP0CZ20}} & \makecell[l]{embeddings \\ performance metrics} & \makecell[l]{PCA \\ line charts} \\ \hline

\makecell[l]{GCN with label \\ propagation \cite{DBLP:journals/corr/abs-2002-06755}} & \makecell[l]{node embeddings \\ performance} & \makecell[l]{graphs \\ line charts} \\ \hline

\bottomrule
\end{tabular}
\end{table*}

Around the same time, an LM visualization survey \cite{DBLP:conf/naacl/LiCHJ16} notes that classic Computer Vision visualization was focused on inverting representations, back-propagation and generating images from sketches, all techniques that work well for images. In NLP, however, it is important to focus on important keywords, composition and dimensional locality; especially since many of the words will depend on the context \cite{DBLP:conf/naacl/LiCHJ16}. Important models will be able to capture this kind of information, and therefore it should be present in visualizations. The early saliency heatmaps clearly showcased these aspects for LSTM and Bi-Directional LSTMs \cite{DBLP:journals/tvcg/StrobeltGPR18}. By later adapting the ideas of first order saliency from Computer Vision, researchers were able to highlight intensification and negation, as well as differences between two sequences at various at consecutive time steps. Their saliency heatmap for SEQ2SEQ auto-encoders also works well for predicting corresponding tokens at each time step \cite{DBLP:conf/acl/LuoJBG19}.

The main characteristic that connects these papers, as it can easily be observed in Table~\ref{tab:explain-rnns}, regardless of the number of visualizations included in them, is the fact that they are focused on a interpretability and explainability of NLP models through visualization. We have analyzed the following characteristics:
\begin{itemize}
    \item \textit{Topic} - the main topic of the paper (e.g., attention, representation, information probing) followed by the papers in which this topic is addressed;
    \item \textit{Visualization Subject} - since visualizations included in these papers were focused on a large set of subjects from Transformer components (e.g., attention heads), to correlation between tasks (e.g., via Pearson correlation charts) or performance (e.g., accuracy or other metrics represented via line charts), we have decided to extract all these in a separate column to understand what kind of charts we might be interested in creating when exploring a certain topic.
    \item \textit{Chart Type} - includes the various types of visual metaphors used for rendering the chosen subjects. Most of the chart types are classic (e.g., line, bar chart, t-SNE), very few being rebranded (e.g., attention maps are heatmaps) or actually new (e.g., comparative attention graphs). The chart names need to give us a clear idea of what they represent.
\end{itemize}

A large class of visualizations is dedicated to the activation of neurons and the representation of hidden states, as it can easily be seen from Table~\ref{tab:explain-rnns}.

LSTMVis \cite{DBLP:journals/tvcg/StrobeltGPR18} uses a large grid of sentences for matching the various state patterns that are then expanded through additional views in the same interface. The key innovation of visualizing hidden state changes keeps the focus on the right keywords, whereas the match views help enlighten particular cases. Many other visualizations similar to LSTMVis (e.g., RNNVis \cite{DBLP:conf/ieeevast/MingCZLCSQ17} or ActiVis \cite{DBLP:journals/tvcg/KahngAKC18}) follow the same template: a control panel is used for selecting the phrase or sentence, a middle view is focused on the word clusters or neuron activations, whereas the last view is typically a matrix view with highlighted cells which showcases the important words that are featured in the activated pathways. Such integrated views offer us holistic views of what these models can accomplish, except for the fact that they are rarely focused also on the corpora that was used for training. In time, the visualization designers started to include this information as well, as we will observe in the next section.

We considered that Graph Convolutional Networks are also worth exploring, but since there are entire libraries dedicated to this task which are not necessarily model-agnostic (e.g., PyTorch Geometric \cite{DBLP:journals/corr/abs-1903-02428}) we have limited ourselves to including some papers that offer some classic visualizations that are typically included in such libraries (e.g., node embeddings, performance).

\subsection{Visualizing Transformers for NLP}\label{sec:transformers-nlp}

No other types of neural networks have led to such an increased demand for custom visualizations since the days of the Kohonen's Self-Organizing Maps \cite{DBLP:books/sp/Kohonen95} or Manbelbrot's fractals \cite{DBLP:books/fm/Mandelbrot77} like the Transformers. When selecting Transformer visualization papers, we decided to focus on the most important topics related to interpretability and explanation.  We  have therefore eliminated papers that used only classic charts (e.g., bars, lines, pies). We decided to focus on the works that tried to visualize as many aspects of Transformer models as possible, from attention maps, to structural or informational probing, neural network layers, and multilingualism. 

Many Transformer visualizations are focused on attention. While the attention mechanism is indeed important for the Transformer architecture, and it improves results for NLP tasks, they are not necessarily easy to interpret if sophisticated encoders are used \cite{DBLP:conf/naacl/JainW19}. Due to this fact, it is often not easy to test various explanations by simply modifying the weights and verifying if the outputs are also changed as a result of this. Alternative theories suggest that simply looking at information flows through such models is not enough, and that attention should only be used as explanation if certain conditions are met, e.g., if the weight distributions found via adversarial training do not perform well \cite{DBLP:conf/emnlp/WiegreffeP19}. Since attention is central to Transformer models, many visualizations are rightly focused on this topic. The fact that such visualizations capture the dynamic nature of the output is not necessarily sufficient to consider them explanations. A good explanation needs to highlight the reasoning chain that lead to the particular output. This is the main reason why we have mainly looked for those visualizations that focus on multiple aspects of the network in this work.

\begin{table*}[]
\caption{Articles focused on explaining Transformer topics through visualizations.}
\label{tab:explain-single}
\centering
\begin{tabular}{@{}lll@{}} \toprule \\
        Topic & Visualization Subject & Chart Type  \\ \midrule

a) Attention & & \\ \hline
\hline
%Jain 
\makecell[l]{attention explanation\\ \cite{DBLP:conf/naacl/JainW19} \cite{DBLP:conf/emnlp/WiegreffeP19} } & \makecell[l]{feature importance correlation\\ permutation\\ adversarial attention \\ performance metrics} & \makecell[l]{Kendall rank statistics \\ scatterplots \\ adversarial charts \\ multiple line charts}\\ \hline

%Abnar 
\makecell[l]{attention flow \\ \cite{DBLP:conf/acl/AbnarZ20} \cite{DBLP:journals/tvcg/DeRoseWB21}}  & \makecell[l]{raw attention \\ raw attention map \\ comparative attention flows } & \makecell[l]{attention graph \\ heatmap \\ comparative attention graphs}\\\hline

%Voita 
\makecell[l]{multi-head self-attention \\ \cite{DBLP:conf/acl/VoitaTMST19}} & \makecell[l]{layers \\ attention for rare words\\ dependency scores\\ active heads } & \makecell[l]{importance charts \\ heatmaps \\bar charts \\ line charts }\\
\hline

b) Hidden states & & \\ \hline
\hline
%Song 
intermediate layers \cite{DBLP:journals/corr/abs-2002-04815}& \makecell[l]{intermediate layers clustering} & \makecell[l]{PCA}\\
\hline
%Hao 
\makecell[l]{information interactions \\ interpretation \cite{DBLP:journals/corr/abs-2004-11207}} & \makecell[l]{scoring attention \\ information flow for tokens \\ evaluation accuracy \\ attention heads correlation} & \makecell[l]{heatmap \\ attribution graphs \\ line charts \\ Pearson correlation charts}\\
\hline

%Vig 
\makecell[l]{causal mediation \\ analysis \cite{DBLP:journals/corr/abs-2004-12265}} & \makecell[l]{indirect effects \\ effects comparison \\attention } & \makecell[l]{heatmaps \\ line chart \\ attention heads}\\\hline

\hline
%Voita 
\makecell[l]{evolution of representations \\ \cite{DBLP:conf/emnlp/VoitaST19a}} & \makecell[l]{token changes and influences \\ distances between layers\\ token occurences} & \makecell[l]{line charts \\ line charts \\ t-SNE clustering}\\
\hline

c) Probing & & \\ \hline
\hline
%Tenney 	
\makecell[l]{structural probes \\ \cite{DBLP:conf/acl/TenneyDP19}}  & \makecell[l]{summary statistics\\ layer-wise performance \\ predictions probing} & \makecell[l]{bar chart\\ bar distribution chart \\multiple bar charts}\\
\hline
%Dufter %Egger	
\makecell[l]{multilingual probes\\  \cite{DBLP:journals/corr/abs-2005-00396} \cite{DBLP:conf/conll/EgerDG20}}   & 
\makecell[l]{
probing task \\ positional embeddings \\ performance metrics \\ stability of training size
} & \makecell[l]{PCA \\ cosine similarity matrices \\ line charts \\ bar charts}
 \\ \hline
 
%Voita  
\makecell[l]{information theoretic \\ probing of classifiers \cite{DBLP:journals/corr/abs-2003-12298}} & \makecell[l]{coding components \\ performance metrics} & \makecell[l]{bar charts \\ line charts}\\
\hline

%Gauthier
psycholinguistics tests \cite{DBLP:conf/acl/GauthierHWQL20} & \makecell[l]{comparing predictions \\ performance metrics \\ test content} & \makecell[l]{tables \\ distribution charts \\ line charts}\\\hline

\bottomrule
\end{tabular}
\end{table*}

The recent success of Transformers helped power many NLP tasks to the top of the leaderboards. BERT visualizations have focused on explaining these great results through visualizations, therefore highlighting: (i) the role of embeddings and relational dependencies within the Transformer learning processes \cite{DBLP:conf/nips/ReifYWVCPK19}; (ii) the role of attention during pre-training or training (e.g., \cite{DBLP:conf/iclr/SuZCLLWD20} or \cite{DBLP:conf/acl/Vig19}) or  (iii) the importance of various linguistic phenomena encoded in its language model like direct objects, noun modifiers, or possessive pronouns. \cite{DBLP:journals/corr/abs-1906-04341}.

Current XAI methods for Transformer models have further developed and supported the idea that understanding the linguistic information which is encoded in the resulting models is key towards understanding the good performances in NLP tasks. Probing tasks \cite{DBLP:conf/acl/BaroniBLKC18} are simple classification problems focused on linguistic features designed to help explore embeddings and LMs. For example, by using structural probing \cite{DBLP:conf/naacl/HewittM19}, structured perceptron parsers \cite{DBLP:journals/corr/abs-2005-01641}) or visualization (e.g., as demonstrated through BERT embeddings and attention layers visualizations like those from \cite{DBLP:journals/corr/abs-1906-04341} and \cite{DBLP:conf/acl/Vig19}), one should be able to understand what kind of linguistic information is encoded into a Transformer model, but also what has changed since previous runs. 

We have discovered two large classes of Transformer visualizations:
\begin{itemize}
    \item \textit{Focused} - visualizations centered on a single subject like attention. The papers themselves might present multiple visualizations, but these visualizations are not single tools.
    \item \textit{Holistic} - visualizations or systems which seek to explain the entire Transformer model or lifecycle.
\end{itemize}

\subsubsection{Focused Transformer Visualizations}
The most important papers dedicated to focused visualizations are summarized in Table~\ref{tab:explain-single}. We used the same conventions in this table like the ones applied in Table~\ref{tab:explain-rnns}.

We can clearly distinguish several large topics in this group of focused papers: the relation between attention and model outputs (e.g., especially in \cite{DBLP:conf/naacl/JainW19}, \cite{DBLP:conf/emnlp/WiegreffeP19}, \cite{DBLP:conf/acl/AbnarZ20}, \cite{DBLP:conf/acl/VoitaTMST19}), the analysis of captured linguistic information via probing (e.g., in \cite{DBLP:conf/emnlp/VoitaST19a},\cite{DBLP:conf/acl/TenneyDP19}), the interpretation of information interaction (e.g., in \cite{DBLP:journals/corr/abs-2004-11207}, \cite{DBLP:conf/emnlp/VoitaST19a}), and multilingualism (e.g., in \cite{DBLP:conf/acl/TenneyDP19} or \cite{DBLP:conf/conll/EgerDG20}). In fact, if we look close to Table~\ref{tab:explain-rnns} we can distinguish 3 large classes of subjects: a) attention; b) hidden states and c) structural or information probing. Papers that work on similar topics also tend to use the same kind of visual metaphors. This sometimes happens due to replication of a previous study (e.g., \cite{DBLP:conf/emnlp/WiegreffeP19} replicates the experiments from \cite{DBLP:conf/naacl/JainW19} to prove that attention weights do not explain everything), whereas in other cases this happens because there is no need for more complicated visual metaphors (e.g., line charts are used in more than half of the papers to represent performance). Besides the widespread use of the heatmaps that represent attention maps, one chart type that deserves to be highlighted in this category is the attention graph \cite{DBLP:journals/corr/abs-2004-11207} which tracks the information flow between the input tokens for a given prediction.

One of the key methods used for explaining the output of LMs is called probing \cite{DBLP:conf/repeval/EttingerER16}, used to analyze the linguistic information encoded in a fixed length vector. Recent iterations like structural probes \cite{DBLP:conf/naacl/HewittM19} evaluate if syntax trees are embedded in a network’s word representation space. Identifying linear transformations is evidence for entire syntax trees embedded in the LM’s vector geometry. This method works well for limited cases in which distances between words are known. The critics of probing argue that differences in probing accuracy between the various classifiers essentially render them unusable as they fail to distinguish between different representations, e.g., two LMs can end up having different linguistic representations even if based on the same initial BERT model. In such scenarios, one cannot compare the accuracy of the classifiers used to predict their labels. To counteract for such cases, a recent information-theoretic probing with minimum description length method was proposed \cite{DBLP:conf/emnlp/VoitaT20}. The basic idea is that instead of predicting labels, the probe transmits data (a description) which is then evaluated based on the returned description’s length. Such probes can be implemented on top of the classic structural probes, and are fairly stable. According to Pilault \cite{DBLP:journals/corr/abs-2002-09084}, classic structural probes may not be enough for complex tasks like summarization simply because it seems that increasing the number of random encoders provide significant performance boosts. This suggests that the information-theoretic probes with minimum description length may be the better probes for this task, however this remains to be demonstrated by future experiments, as Voita’s model was published after Pilault’s article.

\subsubsection{Holistic Transformer Visualizations}

While in Section~\ref{sec:rnns-nlp} several groups tried to expand their visualizations of hidden states to encompass the entirety of the model, they have rarely included the corpora provenance or an easy method to navigate it. Due to this aspect, in the case of Transformers, since such visualizations often included the whole lifecycle (e.g., including the corpora with known provenance), we decided to present them in a separate subsection.

Some of the most interesting tools or papers included in the category of \textit{holistic visualizations} are compared in Table~\ref{tab:explain}. These visualization systems typically integrate most of the components of a Transformer and provide detailed summaries of them. We have examined two large classes of attributes:
\begin{itemize}
    \item \textit{Components} represents the various components of the neural networks: from corpus, to embeddings, positional heads, attention maps or outputs. 
    \item \textit{Summary} includes the various types of views that offer us information about the state of a neuron or a layer, as well as overviews, statistics or details about the various errors encountered. Statistics might include different types of information: from correlations between \textit{layers} or \textit{neurons} to \textit{statistical analyses of the results}. The \textit{errors} column represents any error analysis method through which we can highlight where a particular error comes from (e.g., corpus, training procedure, layer, etc). While it can be argued that neuron or layer views should be included in the components section, the way these views are currently implemented suggests they are rather summaries, as neurons or layers can have different states.
\end{itemize}

We have decided against including chart types in Table~\ref{tab:explain}, as each visualization suite or paper included some novel visualization types besides attention maps (heatmaps), parallel coordinates or line and bar charts.

\begin{table*}[]
\caption{Comparison of holistic Transformer visualizations. Legend: Corpus (C); Embeddings (E); Heads (H); Attention Maps (A); outputs (out); errors (err); neuron (neu); layers (lay); overview (ove); statistics(sta)}.
\label{tab:explain}
\centering
\begin{tabular}{@{}lllllllllll@{}} \toprule
        Article & \multicolumn{4}{l}{Components}
         & \multicolumn{6}{l}{Summary}  \\
        & C & E & H & A & out & err & neu & lay & ove & sta   \\ \midrule

BertViz  \cite{DBLP:journals/corr/abs-1904-02679} &  & \cmark & \cmark & \cmark &  &  & \cmark & \cmark & \cmark & \cmark \\
Clark \cite{DBLP:journals/corr/abs-1906-04341} &  & \cmark & \cmark & \cmark &  & \cmark & \cmark & \cmark & \cmark & \cmark \\
VisBERT \cite{DBLP:conf/www/AkenWLG20}  &  & \cmark & \cmark & \cmark &  &  & \cmark & \cmark & \cmark & \cmark \\
ExBERT \cite{DBLP:conf/acl/HooverSG20} & \cmark & \cmark & \cmark & \cmark & \cmark &  &  & \cmark & \cmark & \cmark \\ 
AttViz \cite{DBLP:journals/corr/abs-2005-05716} & \cmark & \cmark & \cmark & \cmark & \cmark &  &  & \cmark & \cmark & \cmark \\ 
Vector Norms \cite{DBLP:journals/corr/abs-2004-10102}
&  & \cmark & \cmark & \cmark &  &  &  & \cmark & \cmark & \cmark \\
Dictionary Learning \cite{DBLP:journals/corr/abs-2103-15949} 
&  & \cmark & \cmark & \cmark &  &  &  & \cmark & \cmark & \cmark \\

\bottomrule
\end{tabular}
\end{table*}

In our view, none of the examined visualization systems has yet managed to examine all the facets of the Transformers. This is perhaps because this area is relatively new and there is no consensus on what needs to be visualized. While it is quite obvious that individual neurons or attention maps (regardless of if they are averaged or not) are useful, and it is best to visualize them, the same cannot be said about the training corpora today, as only a few systems considered this aspect (e.g., \cite{DBLP:journals/corr/abs-2005-05716} and \cite{DBLP:conf/acl/HooverSG20}). This is not really ideal, as lots of errors might simply come from a bad corpora, but researchers might simply not be aware of them \cite{DBLP:conf/lrec/00020KWN18}. Errors themselves are only seriously discussed in a single publication \cite{DBLP:journals/corr/abs-1906-04341}. ExBERT \cite{DBLP:conf/acl/HooverSG20} and AttViz \cite{DBLP:journals/corr/abs-2005-05716} deserve a special mention here, as they combine different views on the corpus, embeddings and attention maps to provide a holistic image of a Transformer model.

A study that looks at the similarity and stability of neural representations in LMs and brains \cite{DBLP:conf/aaai/HeijdenAS20} shows that combining predictive modelling with Representation Similarity Analysis (RSA) techniques can yield promising results. This article deserves a special mention as it can be included in both focused and holistic visualizations. Their visualizations are rather basic in terms of design, but they contain lots of insights, for example one of the tables they produced showcases the RSA results for various layers of multiple models like BERT, Elmo and others. These kinds of analyses are rather new, and we hope they will become more common in the next years, as they might help us clarify which LMs are more similar to the human brains.

\subsubsection{Vision Transformers}

Transformers have been applied in numerous fields due to their flexibility. Perhaps the most important application is the unification of vision and NLP. In the last two years, this has seen explosive growth. It would be impossible to cover all the various models in this article, and therefore we have selected only a few interesting topics related to this expansion. The one that comes to mind first is Visual Question Answering (VQA) \cite{DBLP:conf/nips/Gan0LZ0020}, as this is perhaps the task that multimodal researchers have been trying to solve for decades. Vision Transformers offer an elegant, but expensive solution. Table~\ref{tab:explain-lav} provides a short summary of interesting papers in this research direction.

\begin{table*}[]
\caption{Articles focused on explaining Language and Vision (LAV) models through visualizations.}
\label{tab:explain-lav}
\centering
\begin{tabular}{@{}lll@{}} \toprule \\
        Domain & Topic & Chart Type  \\ \midrule

%Cao
\makecell[l]{language and vision \\ \cite{DBLP:conf/eccv/CaoGCY0020} \cite{DBLP:journals/corr/abs-2103-00112}} & \makecell[l]{modality importance \\ layer-level importance} & \makecell[l]{attention heatmaps \\ line charts \\ feature maps}\\\hline

%Li
biomedical \cite{DBLP:conf/bibm/LiW020} & attention maps & head attention maps  \\\hline

%Gan
\makecell[l]{VQA \\ \cite{DBLP:conf/nips/Gan0LZ0020} \cite{DBLP:conf/icml/KimSK21}} & \makecell[l]{ multimodal alignment \\ training curves} & \makecell[l]{visual alignment charts \\ line charts} \\\hline

\end{tabular}
\end{table*}

\section{Discussion}\label{sec:discussion}

Multilingual models based on Transformer architectures are error-prone since they are trained on large collections of texts. Due to this training process, the extraction of semantics from language data often captures implicit biases hidden in languages themselves or contextual biases triggered by adaptation to various niches \cite{DBLP:journals/corr/abs-1909-13705}. 

Besides the obvious questions of interpretability (e.g., which features are most important for the prediction, what are best-performing models for creating ensembles?) and explainability (e.g., what linguistic information is actually encoded in the model and why do random encoders perform better for summarization?), another important question is security (e.g., are these models stable enough / do they obtain the same result in any circumstance?). Through adversarial training (Wallace, 2019) it is possible to remove some procedural biases from model-agnostic explainability libraries like LIME or SHAP, as well as well-known attack vectors (e.g., trigger words that might lead to a different language model output) \cite{DBLP:conf/aies/SlackHJSL20}. It can be argued that probing tasks are really only good for simple tasks like NER or POS and for verifying if certain structures were encoded in a LM, but not for more complex tasks like summarization or word-level polysemy disambiguation \cite{DBLP:journals/corr/abs-2103-15949}. Dictionary activation is a method that uses visualization heavily to peek into the activation of words and their senses. Such a method helps uncover both the various layers on which certain word senses are actually learned, and the contexts that might trigger these new senses.

There are several available options  for understanding the inner workings, as well as the results produced by Transformer networks. Each of them have their advantages or shortcomings, briefly discusses in the following.

Model-agnostic tools like the XAI libraries or the hyperparameter optimization and benchmarking tools can be used with a variety of networks. Due to this, model-agnosticism the visualization skills learned while debugging a certain network (e.g., a Convolutional Neural Network) will be easily transferred to debugging and optimizing other networks (e.g., Recurrent Neural Networks). By building a transferable set of skills, users might be more reluctant to try model-specific approaches, like those from the second category discussed in this paper. Some of these model-agnostic tools might be more susceptible to various adversarial attacks \cite{DBLP:journals/corr/abs-1911-02508}, whereas some other tools might not provide us with sufficiently advanced visualizations to match our needs. If the users are already comfortable with some of these options, then they might well be their Swiss-Army knife for any scenario, whereas if they will need specific visualization scenarios (e.g., visualize a specific attention map), it is possible that they will eventually use the Transformer focused visualizations.

While visualization of RNNs started quite early in the DL era, it can be seen as a precursor to the Transformer visualization. Quite often, many of the topics that were important during this ear have remained important during the Transformer era (e.g., hidden states)

Some of the most useful tools discovered during this exploration include: visualization of attention maps (e.g.,  \cite{DBLP:journals/corr/abs-1906-04341}) and embeddings \cite{DBLP:conf/acl/Vig19}, hidden states visualization (e.g., \cite{DBLP:journals/tvcg/StrobeltGPR18}), parallel coordinates plots \cite{DBLP:journals/corr/abs-1906-04341},  and the inclusion of corpus views from ExBERT \cite{DBLP:conf/acl/HooverSG20}.

Current generation of pre-trained LMs based on Transformers \cite{DBLP:journals/corr/abs-1910-03771} was shown to be relatively good at picking up syntactic cues like noun modifiers, possessive pronouns, prepositions or co-referents \cite{DBLP:journals/corr/abs-1906-04341} and semantic cues like entities and relations \cite{DBLP:conf/emnlp/HanGYYLS19}, but has not performed well at capturing different perspectives \cite{DBLP:conf/naacl/ChenK0CR19}, global context \cite{DBLP:conf/emnlp/WaddenWLH19} or relation extraction \cite{DBLP:conf/emnlp/GaoHZLLSZ19}. This may be because biases can be already included in the embeddings and later propagated to the downstream tasks \cite{DBLP:conf/acl-wnlp/GonenG19}.

The MIT Computation Psycholinguistics Laboratory created two useful tools for exploring LMs: the LM Zoo and the SyntaxGym \cite{DBLP:conf/acl/GauthierHWQL20}. The first allows users to install some classic LMs, whereas the SyntaxGym allows users to run pyscholinguistics tests and generate useful visualizations based on their results.

The two large classes of Transformer visualizations we examined (focused on explaining Transformer topics or holistic) are proof that the field is extremely dynamic. While many of the articles focused on explaining Transformer topics like attention or information probing tend to use classic statistical chart types (e.g., bar charts, line charts, PCA, or Pearson correlation charts), we do not consider this a bad thing as we are still in the exploration phase of this technology. Some of these articles also showcase new charts like attention graphs or attention maps. 

The second class of visualizations includes tools like BertViz \cite{DBLP:conf/acl/Vig19}, AttViz \cite{DBLP:journals/corr/abs-2005-05716}, VisBERT \cite{DBLP:conf/www/AkenWLG20} or ExBERT \cite{DBLP:conf/acl/HooverSG20}, that aim to visualize the entire lifecycle of a neural network from corpora and inputs to the model outputs mainly through following the information flow through the various components. They also offer detailed statistics for neurons or network layers. Since most of the models included in this category are rather new, it is expected that this class will expand in the next years.

One important thing to note about visualization methods is that they can easily be imported into other domains. The averaged attention heatmaps used by Vig in his causal mediation analysis for NLP \cite{DBLP:journals/corr/abs-2004-12265}, for example, were later reused for protein analysis in biology \cite{DBLP:journals/corr/abs-2006-15222}. Similarly, attention maps \cite{DBLP:conf/acl/Vig19} developed for BERT models are now used in a wide variety of disciplines, from vision and speech to biology or genetics.

The end goal of future visualization frameworks should be to visualize the entire lifecycle of the Transformer models, from inputs and data sources (e.g., training corpora), to embeddings or attention maps, and finally outputs. In the end, errors observed when creating such models can come from a variety of sources: from the text corpora, from some random network layer or even from some external Knowledge Graph that might feed some data into the model. Tracking such errors would be costly without visualizations.

\section{Conclusion}\label{sec:conclusion}

While the current visualizations aspire to be model-agnostic, we think the directions opened by the various Transformer or RNN visualizations are worthy of expanding upon. In fact, since this is a ubiquitous architecture that has also branched from NLP into areas like semantic video processing, natural language understanding (e.g., speech, translation) and generation (e.g., text generation, music generation), the next generation XAI libraries will probably be built upon it. 

Going beyond current visualizations that are model-agnostic, future frameworks will have to provide visualization components that focus on the important Transformer components like corpora, embeddings, attention heads or additional neural network layers that might be problem-specific. By focusing on the common components from larger architectures, it should be possible to enhance reusability. Other important features that should be included in future frameworks are the ability to summarize the model's state (e.g., through averaged attention heatmaps or similar visualization mechanisms) at various levels (e.g., neurons, layers, inputs and outputs), as well as the possibility to compare multiple settings for one or multiple models. It is important to note that most of the current visualizations seem to be designed to showcase methods and properties that were already known to belong to neural networks. It would be interesting to design visualizations that allow us to explore neural networks, that help us discover new properties. Such interactive exploration tools would significantly expand the role of visualization from communication to knowledge discovery. 

One interesting direction is the automated development of model specific visualizations, as more complex neural networks might also include many specific components that cannot always be included into more general model agnostic frameworks. 

\bibliographystyle{unsrtnat}
\bibliography{visualizing-explaining-language-models.bib}  %%% Uncomment this line and comment out the ``thebibliography'' section below to use the external .bib file (using bibtex) .

\begin{thebibliography}{108}
\providecommand{\natexlab}[1]{#1}
\providecommand{\url}[1]{\texttt{#1}}
\expandafter\ifx\csname urlstyle\endcsname\relax
  \providecommand{\doi}[1]{doi: #1}\else
  \providecommand{\doi}{doi: \begingroup \urlstyle{rm}\Url}\fi

\bibitem[Ponte and Croft(1998)]{DBLP:conf/sigir/PonteC98}
Jay~M. Ponte and W.~Bruce Croft.
\newblock A language modeling approach to information retrieval.
\newblock In W.~Bruce Croft, Alistair Moffat, C.~J. van Rijsbergen, Ross
  Wilkinson, and Justin Zobel, editors, \emph{{SIGIR} '98: Proceedings of the
  21st Annual International {ACM} {SIGIR} Conference on Research and
  Development in Information Retrieval, August 24-28 1998, Melbourne,
  Australia}, pages 275--281. {ACM}, 1998.
\newblock ISBN 1-58113-015-5.
\newblock \doi{10.1145/290941.291008}.
\newblock URL \url{https://doi.org/10.1145/290941.291008}.

\bibitem[Brown et~al.(2020)Brown, Mann, Ryder, Subbiah, Kaplan, Dhariwal,
  Neelakantan, Shyam, Sastry, Askell, Agarwal, Herbert{-}Voss, Krueger,
  Henighan, Child, Ramesh, Ziegler, Wu, Winter, Hesse, Chen, Sigler, Litwin,
  Gray, Chess, Clark, Berner, McCandlish, Radford, Sutskever, and
  Amodei]{DBLP:conf/nips/BrownMRSKDNSSAA20}
Tom~B. Brown, Benjamin Mann, Nick Ryder, Melanie Subbiah, Jared Kaplan,
  Prafulla Dhariwal, Arvind Neelakantan, Pranav Shyam, Girish Sastry, Amanda
  Askell, Sandhini Agarwal, Ariel Herbert{-}Voss, Gretchen Krueger, Tom
  Henighan, Rewon Child, Aditya Ramesh, Daniel~M. Ziegler, Jeffrey Wu, Clemens
  Winter, Christopher Hesse, Mark Chen, Eric Sigler, Mateusz Litwin, Scott
  Gray, Benjamin Chess, Jack Clark, Christopher Berner, Sam McCandlish, Alec
  Radford, Ilya Sutskever, and Dario Amodei.
\newblock Language models are few-shot learners.
\newblock In Hugo Larochelle, Marc'Aurelio Ranzato, Raia Hadsell,
  Maria{-}Florina Balcan, and Hsuan{-}Tien Lin, editors, \emph{Advances in
  Neural Information Processing Systems 33: Annual Conference on Neural
  Information Processing Systems 2020, NeurIPS 2020, December 6-12, 2020,
  virtual}, 2020.
\newblock URL
  \url{https://proceedings.neurips.cc/paper/2020/hash/1457c0d6bfcb496741-8bfb8ac142f64a-Abstract.html}.

\bibitem[Vaswani et~al.(2017)Vaswani, Shazeer, Parmar, Uszkoreit, Jones, Gomez,
  Kaiser, and Polosukhin]{DBLP:conf/nips/VaswaniSPUJGKP17}
Ashish Vaswani, Noam Shazeer, Niki Parmar, Jakob Uszkoreit, Llion Jones,
  Aidan~N. Gomez, Lukasz Kaiser, and Illia Polosukhin.
\newblock Attention is all you need.
\newblock In Isabelle Guyon, Ulrike von Luxburg, Samy Bengio, Hanna~M. Wallach,
  Rob Fergus, S.~V.~N. Vishwanathan, and Roman Garnett, editors, \emph{Advances
  in Neural Information Processing Systems 30: Annual Conference on Neural
  Information Processing Systems 2017, December 4-9, 2017, Long Beach, CA,
  {USA}}, pages 5998--6008, 2017.
\newblock URL
  \url{https://proceedings.neurips.cc/paper/2017/hash/3f5ee243547dee91fbd053c1c4a845aa-Abstract.html}.

\bibitem[Raghu and Schmidt(2020)]{DBLP:journals/corr/abs-2003-11755}
Maithra Raghu and Eric Schmidt.
\newblock A survey of deep learning for scientific discovery.
\newblock \emph{CoRR}, abs/2003.11755, 2020.
\newblock URL \url{https://arxiv.org/abs/2003.11755}.

\bibitem[Devlin et~al.(2019)Devlin, Chang, Lee, and
  Toutanova]{DBLP:conf/naacl/DevlinCLT19}
Jacob Devlin, Ming{-}Wei Chang, Kenton Lee, and Kristina Toutanova.
\newblock {BERT:} pre-training of deep bidirectional transformers for language
  understanding.
\newblock In Jill Burstein, Christy Doran, and Thamar Solorio, editors,
  \emph{Proceedings of the 2019 Conference of the North American Chapter of the
  Association for Computational Linguistics: Human Language Technologies,
  {NAACL-HLT} 2019, Minneapolis, MN, USA, June 2-7, 2019, Volume 1 (Long and
  Short Papers)}, pages 4171--4186. Association for Computational Linguistics,
  2019.
\newblock \doi{10.18653/v1/n19-1423}.
\newblock URL \url{https://doi.org/10.18653/v1/n19-1423}.

\bibitem[Liu et~al.(2019)Liu, Ott, Goyal, Du, Joshi, Chen, Levy, Lewis,
  Zettlemoyer, and Stoyanov]{DBLP:journals/corr/abs-1907-11692}
Yinhan Liu, Myle Ott, Naman Goyal, Jingfei Du, Mandar Joshi, Danqi Chen, Omer
  Levy, Mike Lewis, Luke Zettlemoyer, and Veselin Stoyanov.
\newblock Roberta: {A} robustly optimized {BERT} pretraining approach.
\newblock \emph{CoRR}, abs/1907.11692, 2019.
\newblock URL \url{http://arxiv.org/abs/1907.11692}.

\bibitem[Lan et~al.(2020)Lan, Chen, Goodman, Gimpel, Sharma, and
  Soricut]{DBLP:conf/iclr/LanCGGSS20}
Zhenzhong Lan, Mingda Chen, Sebastian Goodman, Kevin Gimpel, Piyush Sharma, and
  Radu Soricut.
\newblock {ALBERT:} {A} lite {BERT} for self-supervised learning of language
  representations.
\newblock In \emph{8th International Conference on Learning Representations,
  {ICLR} 2020, Addis Ababa, Ethiopia, April 26-30, 2020}. OpenReview.net, 2020.
\newblock URL \url{https://openreview.net/forum?id=H1eA7AEtvS}.

\bibitem[Yang et~al.(2019)Yang, Dai, Yang, Carbonell, Salakhutdinov, and
  Le]{DBLP:conf/nips/YangDYCSL19}
Zhilin Yang, Zihang Dai, Yiming Yang, Jaime~G. Carbonell, Ruslan Salakhutdinov,
  and Quoc~V. Le.
\newblock Xlnet: Generalized autoregressive pretraining for language
  understanding.
\newblock In Hanna~M. Wallach, Hugo Larochelle, Alina Beygelzimer, Florence
  d'Alch{\'{e}}{-}Buc, Emily~B. Fox, and Roman Garnett, editors, \emph{Advances
  in Neural Information Processing Systems 32: Annual Conference on Neural
  Information Processing Systems 2019, NeurIPS 2019, December 8-14, 2019,
  Vancouver, BC, Canada}, pages 5754--5764, 2019.
\newblock URL
  \url{https://proceedings.neurips.cc/paper/2019/hash/dc6a7e655d7e5840e6-6733e9ee67cc69-Abstract.html}.

\bibitem[Sanh et~al.(2019)Sanh, Debut, Chaumond, and
  Wolf]{DBLP:journals/corr/abs-1910-01108}
Victor Sanh, Lysandre Debut, Julien Chaumond, and Thomas Wolf.
\newblock Distilbert, a distilled version of {BERT:} smaller, faster, cheaper
  and lighter.
\newblock \emph{CoRR}, abs/1910.01108, 2019.
\newblock URL \url{http://arxiv.org/abs/1910.01108}.

\bibitem[Kitaev et~al.(2020)Kitaev, Kaiser, and
  Levskaya]{DBLP:conf/iclr/KitaevKL20}
Nikita Kitaev, Lukasz Kaiser, and Anselm Levskaya.
\newblock Reformer: The efficient transformer.
\newblock In \emph{8th International Conference on Learning Representations,
  {ICLR} 2020, Addis Ababa, Ethiopia, April 26-30, 2020}. OpenReview.net, 2020.
\newblock URL \url{https://openreview.net/forum?id=rkgNKkHtvB}.

\bibitem[Wolf et~al.(2019)Wolf, Debut, Sanh, Chaumond, Delangue, Moi, Cistac,
  Rault, Louf, Funtowicz, and Brew]{DBLP:journals/corr/abs-1910-03771}
Thomas Wolf, Lysandre Debut, Victor Sanh, Julien Chaumond, Clement Delangue,
  Anthony Moi, Pierric Cistac, Tim Rault, R{\'{e}}mi Louf, Morgan Funtowicz,
  and Jamie Brew.
\newblock Huggingface transformers: State-of-the-art natural language
  processing.
\newblock \emph{CoRR}, abs/1910.03771, 2019.
\newblock URL \url{http://arxiv.org/abs/1910.03771}.

\bibitem[Bra{\c{s}}oveanu and Andonie(2020)]{BrasoveanuA20}
Adrian~MP Bra{\c{s}}oveanu and R{\u{a}}zvan Andonie.
\newblock Visualizing transformers for nlp: a brief survey.
\newblock In \emph{2020 24th International Conference Information Visualisation
  (IV)}, pages 270--279. IEEE, 2020.
\newblock ISBN 978-1-7281-9134-8.
\newblock \doi{10.1109/IV51561.2020}.

\bibitem[Wilkinson(2005)]{DBLP:books/daglib/0024564}
Leland Wilkinson.
\newblock \emph{The Grammar of Graphics, Second Edition}.
\newblock Statistics and computing. Springer, 2005.
\newblock ISBN 978-0-387-24544-7.

\bibitem[Bostock et~al.(2011)Bostock, Ogievetsky, and
  Heer]{DBLP:journals/tvcg/BostockOH11}
Michael Bostock, Vadim Ogievetsky, and Jeffrey Heer.
\newblock D{\({^3}\)} data-driven documents.
\newblock \emph{{IEEE} Trans. Vis. Comput. Graph.}, 17\penalty0 (12):\penalty0
  2301--2309, 2011.
\newblock \doi{10.1109/TVCG.2011.185}.
\newblock URL \url{https://doi.org/10.1109/TVCG.2011.185}.

\bibitem[Satyanarayan et~al.(2017)Satyanarayan, Moritz, Wongsuphasawat, and
  Heer]{DBLP:journals/tvcg/SatyanarayanMWH17}
Arvind Satyanarayan, Dominik Moritz, Kanit Wongsuphasawat, and Jeffrey Heer.
\newblock Vega-lite: {A} grammar of interactive graphics.
\newblock \emph{{IEEE} Trans. Vis. Comput. Graph.}, 23\penalty0 (1):\penalty0
  341--350, 2017.
\newblock \doi{10.1109/TVCG.2016.2599030}.
\newblock URL \url{https://doi.org/10.1109/TVCG.2016.2599030}.

\bibitem[Park et~al.(2018)Park, Kim, Lee, Choo, Diakopoulos, and
  Elmqvist]{DBLP:journals/tvcg/ParkKLCDE18}
Deokgun Park, Seungyeon Kim, Jurim Lee, Jaegul Choo, Nicholas Diakopoulos, and
  Niklas Elmqvist.
\newblock Conceptvector: Text visual analytics via interactive lexicon building
  using word embedding.
\newblock \emph{{IEEE} Trans. Vis. Comput. Graph.}, 24\penalty0 (1):\penalty0
  361--370, 2018.
\newblock \doi{10.1109/TVCG.2017.2744478}.
\newblock URL \url{https://doi.org/10.1109/TVCG.2017.2744478}.

\bibitem[Hoover et~al.(2020)Hoover, Strobelt, and
  Gehrmann]{DBLP:conf/acl/HooverSG20}
Benjamin Hoover, Hendrik Strobelt, and Sebastian Gehrmann.
\newblock ex bert: {A} visual analysis tool to explore learned representations
  in transformer models.
\newblock In Asli Celikyilmaz and Tsung{-}Hsien Wen, editors, \emph{Proceedings
  of the 58th Annual Meeting of the Association for Computational Linguistics:
  System Demonstrations, {ACL} 2020, Online, July 5-10, 2020}, pages 187--196.
  Association for Computational Linguistics, 2020.
\newblock \doi{10.18653/v1/2020.acl-demos.22}.
\newblock URL \url{https://doi.org/10.18653/v1/2020.acl-demos.22}.

\bibitem[Vig(2019{\natexlab{a}})]{DBLP:conf/acl/Vig19}
Jesse Vig.
\newblock A multiscale visualization of attention in the transformer model.
\newblock In Marta~R. Costa{-}juss{\`{a}} and Enrique Alfonseca, editors,
  \emph{Proceedings of the 57th Conference of the Association for Computational
  Linguistics, {ACL} 2019, Florence, Italy, July 28 - August 2, 2019, Volume 3:
  System Demonstrations}, pages 37--42. Association for Computational
  Linguistics, 2019{\natexlab{a}}.
\newblock ISBN 978-1-950737-49-9.
\newblock \doi{10.18653/v1/p19-3007}.
\newblock URL \url{https://doi.org/10.18653/v1/p19-3007}.

\bibitem[Clark et~al.(2019)Clark, Khandelwal, Levy, and
  Manning]{DBLP:journals/corr/abs-1906-04341}
Kevin Clark, Urvashi Khandelwal, Omer Levy, and Christopher~D. Manning.
\newblock What does {BERT} look at? an analysis of bert's attention.
\newblock \emph{CoRR}, abs/1906.04341, 2019.
\newblock URL \url{http://arxiv.org/abs/1906.04341}.

\bibitem[Florea and Andonie(2019)]{DBLP:journals/ijccc/FloreaA19}
Adrian Florea and Razvan Andonie.
\newblock Weighted random search for hyperparameter optimization.
\newblock \emph{Int. J. Comput. Commun. Control}, 14\penalty0 (2):\penalty0
  154--169, 2019.
\newblock \doi{10.15837/ijccc.2019.2.3514}.
\newblock URL \url{https://doi.org/10.15837/ijccc.2019.2.3514}.

\bibitem[Heinrich and Weiskopf(2015)]{DBLP:journals/cse/HeinrichW15}
Julian Heinrich and Daniel Weiskopf.
\newblock Parallel coordinates for multidimensional data visualization: Basic
  concepts.
\newblock \emph{Comput. Sci. Eng.}, 17\penalty0 (3):\penalty0 70--76, 2015.
\newblock \doi{10.1109/MCSE.2015.55}.
\newblock URL \url{https://doi.org/10.1109/MCSE.2015.55}.

\bibitem[Moritz et~al.(2018)Moritz, Nishihara, Wang, Tumanov, Liaw, Liang,
  Elibol, Yang, Paul, Jordan, and Stoica]{DBLP:conf/osdi/MoritzNWTLLEYPJ18}
Philipp Moritz, Robert Nishihara, Stephanie Wang, Alexey Tumanov, Richard Liaw,
  Eric Liang, Melih Elibol, Zongheng Yang, William Paul, Michael~I. Jordan, and
  Ion Stoica.
\newblock Ray: {A} distributed framework for emerging {AI} applications.
\newblock In Andrea~C. Arpaci{-}Dusseau and Geoff Voelker, editors, \emph{13th
  {USENIX} Symposium on Operating Systems Design and Implementation, {OSDI}
  2018, Carlsbad, CA, USA, October 8-10, 2018}, pages 561--577. {USENIX}
  Association, 2018.
\newblock URL
  \url{https://www.usenix.org/conference/osdi18/presentation/nishihara}.

\bibitem[Liaw et~al.(2018)Liaw, Liang, Nishihara, Moritz, Gonzalez, and
  Stoica]{DBLP:journals/corr/abs-1807-05118}
Richard Liaw, Eric Liang, Robert Nishihara, Philipp Moritz, Joseph~E. Gonzalez,
  and Ion Stoica.
\newblock Tune: {A} research platform for distributed model selection and
  training.
\newblock \emph{CoRR}, abs/1807.05118, 2018.
\newblock URL \url{http://arxiv.org/abs/1807.05118}.

\bibitem[Strubell et~al.(2019)Strubell, Ganesh, and
  McCallum]{DBLP:conf/acl/StrubellGM19}
Emma Strubell, Ananya Ganesh, and Andrew McCallum.
\newblock Energy and policy considerations for deep learning in {NLP}.
\newblock In Anna Korhonen, David~R. Traum, and Lluis Marquez, editors,
  \emph{Proceedings of the 57th Conference of the Association for Computational
  Linguistics, {ACL} 2019, Florence, Italy, July 28- August 2, 2019, Volume 1:
  Long Papers}, pages 3645--3650. Association for Computational Linguistics,
  2019.
\newblock \doi{10.18653/v1/p19-1355}.
\newblock URL \url{https://doi.org/10.18653/v1/p19-1355}.

\bibitem[Shwartz and Choi(2020)]{DBLP:conf/coling/ShwartzC20}
Vered Shwartz and Yejin Choi.
\newblock Do neural language models overcome reporting bias?
\newblock In Donia Scott, N{\'{u}}ria Bel, and Chengqing Zong, editors,
  \emph{Proceedings of the 28th International Conference on Computational
  Linguistics, {COLING} 2020, Barcelona, Spain (Online), December 8-13, 2020},
  pages 6863--6870. International Committee on Computational Linguistics, 2020.
\newblock \doi{10.18653/v1/2020.coling-main.605}.
\newblock URL \url{https://doi.org/10.18653/v1/2020.coling-main.605}.

\bibitem[Vellido(2020)]{DBLP:journals/nca/Vellido20}
Alfredo Vellido.
\newblock The importance of interpretability and visualization in machine
  learning for applications in medicine and health care.
\newblock \emph{Neural Comput. Appl.}, 32\penalty0 (24):\penalty0 18069--18083,
  2020.
\newblock \doi{10.1007/s00521-019-04051-w}.
\newblock URL \url{https://doi.org/10.1007/s00521-019-04051-w}.

\bibitem[Heer(2019)]{DBLP:journals/pnas/Heer19}
Jeffrey Heer.
\newblock Agency plus automation: Designing artificial intelligence into
  interactive systems.
\newblock \emph{Proc. Natl. Acad. Sci. {USA}}, 116\penalty0 (6):\penalty0
  1844--1850, 2019.
\newblock \doi{10.1073/pnas.1807184115}.
\newblock URL \url{https://doi.org/10.1073/pnas.1807184115}.

\bibitem[Qin et~al.(2018)Qin, Yu, Liu, and Chen]{DBLP:journals/mfc/QinYLC18}
Zhuwei Qin, Fuxun Yu, Chenchen Liu, and Xiang Chen.
\newblock How convolutional neural networks see the world - {A} survey of
  convolutional neural network visualization methods.
\newblock \emph{Math. Found. Comput.}, 1\penalty0 (2):\penalty0 149--180, 2018.
\newblock \doi{10.3934/mfc.2018008}.
\newblock URL \url{https://doi.org/10.3934/mfc.2018008}.

\bibitem[Zhang and Zhu(2018)]{DBLP:journals/jzusc/ZhangZ18}
Quanshi Zhang and Song{-}Chun Zhu.
\newblock Visual interpretability for deep learning: a survey.
\newblock \emph{Frontiers Inf. Technol. Electron. Eng.}, 19\penalty0
  (1):\penalty0 27--39, 2018.
\newblock \doi{10.1631/FITEE.1700808}.
\newblock URL \url{https://doi.org/10.1631/FITEE.1700808}.

\bibitem[Nguyen et~al.(2019)Nguyen, Yosinski, and
  Clune]{DBLP:series/lncs/NguyenYC19}
Anh Nguyen, Jason Yosinski, and Jeff Clune.
\newblock Understanding neural networks via feature visualization: {A} survey.
\newblock In  \citet{DBLP:series/lncs/11700}, pages 55--76.
\newblock ISBN 978-3-030-28953-9.
\newblock \doi{10.1007/978-3-030-28954-6\_4}.
\newblock URL \url{https://doi.org/10.1007/978-3-030-28954-6\_4}.

\bibitem[Hohman et~al.(2019)Hohman, Kahng, Pienta, and
  Chau]{DBLP:journals/tvcg/HohmanKPC19}
Fred Hohman, Minsuk Kahng, Robert Pienta, and Duen~Horng Chau.
\newblock Visual analytics in deep learning: An interrogative survey for the
  next frontiers.
\newblock \emph{{IEEE} Trans. Vis. Comput. Graph.}, 25\penalty0 (8):\penalty0
  2674--2693, 2019.
\newblock \doi{10.1109/TVCG.2018.2843369}.
\newblock URL \url{https://doi.org/10.1109/TVCG.2018.2843369}.

\bibitem[Samek et~al.(2019)Samek, Montavon, Vedaldi, Hansen, and
  M{\"{u}}ller]{DBLP:series/lncs/11700}
Wojciech Samek, Gr{\'{e}}goire Montavon, Andrea Vedaldi, Lars~Kai Hansen, and
  Klaus{-}Robert M{\"{u}}ller, editors.
\newblock \emph{Explainable {AI:} Interpreting, Explaining and Visualizing Deep
  Learning}, volume 11700 of \emph{Lecture Notes in Computer Science}.
\newblock Springer, 2019.
\newblock ISBN 978-3-030-28953-9.
\newblock \doi{10.1007/978-3-030-28954-6}.
\newblock URL \url{https://doi.org/10.1007/978-3-030-28954-6}.

\bibitem[Guyon and Elisseeff(2003)]{DBLP:journals/jmlr/GuyonE03}
Isabelle Guyon and Andr{\'{e}} Elisseeff.
\newblock An introduction to variable and feature selection.
\newblock \emph{J. Mach. Learn. Res.}, 3:\penalty0 1157--1182, 2003.
\newblock URL \url{http://jmlr.org/papers/v3/guyon03a.html}.

\bibitem[Ribeiro et~al.(2016)Ribeiro, Singh, and
  Guestrin]{DBLP:conf/kdd/Ribeiro0G16}
Marco~T{\'{u}}lio Ribeiro, Sameer Singh, and Carlos Guestrin.
\newblock "why should {I} trust you?": Explaining the predictions of any
  classifier.
\newblock In Balaji Krishnapuram, Mohak Shah, Alexander~J. Smola, Charu~C.
  Aggarwal, Dou Shen, and Rajeev Rastogi, editors, \emph{Proceedings of the
  22nd {ACM} {SIGKDD} International Conference on Knowledge Discovery and Data
  Mining, San Francisco, CA, USA, August 13-17, 2016}, pages 1135--1144. {ACM},
  2016.
\newblock ISBN 978-1-4503-4232-2.
\newblock \doi{10.1145/2939672.2939778}.
\newblock URL \url{https://doi.org/10.1145/2939672.2939778}.

\bibitem[Lundberg and Lee(2017)]{DBLP:conf/nips/LundbergL17}
Scott~M. Lundberg and Su{-}In Lee.
\newblock A unified approach to interpreting model predictions.
\newblock In Isabelle Guyon, Ulrike von Luxburg, Samy Bengio, Hanna~M. Wallach,
  Rob Fergus, S.~V.~N. Vishwanathan, and Roman Garnett, editors, \emph{Advances
  in Neural Information Processing Systems 30: Annual Conference on Neural
  Information Processing Systems 2017, December 4-9, 2017, Long Beach, CA,
  {USA}}, pages 4765--4774, 2017.
\newblock URL
  \url{https://proceedings.neurips.cc/paper/2017/hash/8a20a8621978632d76c43dfd28b67767-Abstract.html}.

\bibitem[Fan et~al.(2019)Fan, Jernite, Perez, Grangier, Weston, and
  Auli]{DBLP:conf/acl/FanJPGWA19}
Angela Fan, Yacine Jernite, Ethan Perez, David Grangier, Jason Weston, and
  Michael Auli.
\newblock {ELI5:} long form question answering.
\newblock In Anna Korhonen, David~R. Traum, and Lluis Marquez, editors,
  \emph{Proceedings of the 57th Conference of the Association for Computational
  Linguistics, {ACL} 2019, Florence, Italy, July 28- August 2, 2019, Volume 1:
  Long Papers}, pages 3558--3567. Association for Computational Linguistics,
  2019.
\newblock \doi{10.18653/v1/p19-1346}.
\newblock URL \url{https://doi.org/10.18653/v1/p19-1346}.

\bibitem[Slack et~al.(2020)Slack, Hilgard, Jia, Singh, and
  Lakkaraju]{DBLP:conf/aies/SlackHJSL20}
Dylan Slack, Sophie Hilgard, Emily Jia, Sameer Singh, and Himabindu Lakkaraju.
\newblock Fooling {LIME} and {SHAP:} adversarial attacks on post hoc
  explanation methods.
\newblock In Annette~N. Markham, Julia Powles, Toby Walsh, and Anne~L.
  Washington, editors, \emph{{AIES} '20: {AAAI/ACM} Conference on AI, Ethics,
  and Society, New York, NY, USA, February 7-8, 2020}, pages 180--186. {ACM},
  2020.
\newblock ISBN 978-1-4503-7110-0.
\newblock \doi{10.1145/3375627.3375830}.
\newblock URL \url{https://doi.org/10.1145/3375627.3375830}.

\bibitem[Lundberg et~al.(2018)Lundberg, Nair, Vavilala, Horibe, Eisses, Adams,
  Liston, Low, Newman, Kim, et~al.]{lundberg2018explainable}
Scott~M Lundberg, Bala Nair, Monica~S Vavilala, Mayumi Horibe, Michael~J
  Eisses, Trevor Adams, David~E Liston, Daniel King-Wai Low, Shu-Fang Newman,
  Jerry Kim, et~al.
\newblock Explainable machine-learning predictions for the prevention of
  hypoxaemia during surgery.
\newblock \emph{Nature biomedical engineering}, 2\penalty0 (10):\penalty0
  749--760, 2018.

\bibitem[Wallace et~al.(2019{\natexlab{a}})Wallace, Feng, Kandpal, Gardner, and
  Singh]{DBLP:conf/emnlp/WallaceFKGS19}
Eric Wallace, Shi Feng, Nikhil Kandpal, Matt Gardner, and Sameer Singh.
\newblock Universal adversarial triggers for attacking and analyzing {NLP}.
\newblock In Kentaro Inui, Jing Jiang, Vincent Ng, and Xiaojun Wan, editors,
  \emph{Proceedings of the 2019 Conference on Empirical Methods in Natural
  Language Processing and the 9th International Joint Conference on Natural
  Language Processing, {EMNLP-IJCNLP} 2019, Hong Kong, China, November 3-7,
  2019}, pages 2153--2162. Association for Computational Linguistics,
  2019{\natexlab{a}}.
\newblock \doi{10.18653/v1/D19-1221}.
\newblock URL \url{https://doi.org/10.18653/v1/D19-1221}.

\bibitem[Carlini et~al.(2020)Carlini, Tram{\`{e}}r, Wallace, Jagielski,
  Herbert{-}Voss, Lee, Roberts, Brown, Song, Erlingsson, Oprea, and
  Raffel]{DBLP:journals/corr/abs-2012-07805}
Nicholas Carlini, Florian Tram{\`{e}}r, Eric Wallace, Matthew Jagielski, Ariel
  Herbert{-}Voss, Katherine Lee, Adam Roberts, Tom~B. Brown, Dawn Song,
  {\'{U}}lfar Erlingsson, Alina Oprea, and Colin Raffel.
\newblock Extracting training data from large language models.
\newblock \emph{CoRR}, abs/2012.07805, 2020.
\newblock URL \url{https://arxiv.org/abs/2012.07805}.

\bibitem[Sun et~al.(2020)Sun, Hashimoto, Yin, Asai, Li, Yu, and
  Xiong]{DBLP:journals/corr/abs-2003-04985}
Lichao Sun, Kazuma Hashimoto, Wenpeng Yin, Akari Asai, Jia Li, Philip~S. Yu,
  and Caiming Xiong.
\newblock Adv-bert: {BERT} is not robust on misspellings! generating nature
  adversarial samples on {BERT}.
\newblock \emph{CoRR}, abs/ 2003.04985, 2020.
\newblock URL \url{https://arxiv.org/abs/2003.04985}.

\bibitem[Ebrahimi et~al.(2018)Ebrahimi, Rao, Lowd, and
  Dou]{DBLP:conf/acl/EbrahimiRLD18}
Javid Ebrahimi, Anyi Rao, Daniel Lowd, and Dejing Dou.
\newblock Hotflip: White-box adversarial examples for text classification.
\newblock In Iryna Gurevych and Yusuke Miyao, editors, \emph{Proceedings of the
  56th Annual Meeting of the Association for Computational Linguistics, {ACL}
  2018, Melbourne, Australia, July 15-20, 2018, Volume 2: Short Papers}, pages
  31--36. Association for Computational Linguistics, 2018.
\newblock ISBN 978-1-948087-34-6.
\newblock \doi{10.18653/v1/P18-2006}.
\newblock URL \url{https://www.aclweb.org/anthology/P18-2006/}.

\bibitem[Jin et~al.(2020)Jin, Jin, Zhou, and
  Szolovits]{DBLP:conf/aaai/JinJZS20}
Di~Jin, Zhijing Jin, Joey~Tianyi Zhou, and Peter Szolovits.
\newblock Is {BERT} really robust? {A} strong baseline for natural language
  attack on text classification and entailment.
\newblock In \emph{The Thirty-Fourth {AAAI} Conference on Artificial
  Intelligence, {AAAI} 2020, The Thirty-Second Innovative Applications of
  Artificial Intelligence Conference, {IAAI} 2020, The Tenth {AAAI} Symposium
  on Educational Advances in Artificial Intelligence, {EAAI} 2020, New York,
  NY, USA, February 7-12, 2020}, pages 8018--8025. {AAAI} Press, 2020.
\newblock URL \url{https://aaai.org/ojs/index.php/AAAI/article/view/6311}.

\bibitem[Morris et~al.(2020)Morris, Lifland, Yoo, Grigsby, Jin, and
  Qi]{DBLP:conf/emnlp/MorrisLYGJQ20}
John~X. Morris, Eli Lifland, Jin~Yong Yoo, Jake Grigsby, Di~Jin, and Yanjun Qi.
\newblock Textattack: {A} framework for adversarial attacks, data augmentation,
  and adversarial training in {NLP}.
\newblock In Qun Liu and David Schlangen, editors, \emph{Proceedings of the
  2020 Conference on Empirical Methods in Natural Language Processing: System
  Demonstrations, {EMNLP} 2020 - Demos, Online, November 16-20, 2020}, pages
  119--126. Association for Computational Linguistics, 2020.
\newblock ISBN 978-1-952148-62-0.
\newblock \doi{10.18653/v1/2020.emnlp-demos.16}.
\newblock URL \url{https://doi.org/10.18653/v1/2020.emnlp-demos.16}.

\bibitem[Zeng et~al.(2020)Zeng, Qi, Zhou, Zhang, Hou, Zang, Liu, and
  Sun]{DBLP:journals/corr/abs-2009-09191}
Guoyang Zeng, Fanchao Qi, Qianrui Zhou, Tingji Zhang, Bairu Hou, Yuan Zang,
  Zhiyuan Liu, and Maosong Sun.
\newblock Openattack: An open-source textual adversarial attack toolkit.
\newblock \emph{CoRR}, abs/2009.09191, 2020.
\newblock URL \url{https://arxiv.org/abs/2009.09191}.

\bibitem[Karimi et~al.(2020)Karimi, Rossi, and
  Prati]{DBLP:conf/icpr/KarimiR020}
Akbar Karimi, Leonardo Rossi, and Andrea Prati.
\newblock Adversarial training for aspect-based sentiment analysis with {BERT}.
\newblock In \emph{25th International Conference on Pattern Recognition, {ICPR}
  2020, Virtual Event / Milan, Italy, January 10-15, 2021}, pages 8797--8803.
  {IEEE}, 2020.
\newblock ISBN 978-1-7281-8808-9.
\newblock \doi{10.1109/ICPR48806.2021.9412167}.
\newblock URL \url{https://doi.org/10.1109/ICPR48806.2021.9412167}.

\bibitem[Hao et~al.(2020)Hao, Dong, Wei, and
  Xu]{DBLP:journals/corr/abs-2004-11207}
Yaru Hao, Li~Dong, Furu Wei, and Ke~Xu.
\newblock Self-attention attribution: Interpreting information interactions
  inside transformer.
\newblock \emph{CoRR}, abs/2004.11207, 2020.
\newblock URL \url{https://arxiv.org/abs/2004.11207}.

\bibitem[Chen et~al.(2020)Chen, Salem, Backes, Ma, and
  Zhang]{DBLP:journals/corr/abs-2006-01043}
Xiaoyi Chen, Ahmed Salem, Michael Backes, Shiqing Ma, and Yang Zhang.
\newblock Badnl: Backdoor attacks against {NLP} models.
\newblock \emph{CoRR}, abs/2006.01043, 2020.
\newblock URL \url{https://arxiv.org/abs/2006.01043}.

\bibitem[Wallace et~al.(2019{\natexlab{b}})Wallace, Tuyls, Wang, Subramanian,
  Gardner, and Singh]{DBLP:conf/emnlp/WallaceTWSGS19}
Eric Wallace, Jens Tuyls, Junlin Wang, Sanjay Subramanian, Matt Gardner, and
  Sameer Singh.
\newblock Allennlp interpret: {A} framework for explaining predictions of {NLP}
  models.
\newblock In Sebastian Pad{\'{o}} and Ruihong Huang, editors, \emph{Proceedings
  of the 2019 Conference on Empirical Methods in Natural Language Processing
  and the 9th International Joint Conference on Natural Language Processing,
  {EMNLP-IJCNLP} 2019, Hong Kong, China, November 3-7, 2019 - System
  Demonstrations}, pages 7--12. Association for Computational Linguistics,
  2019{\natexlab{b}}.
\newblock \doi{10.18653/v1/D19-3002}.
\newblock URL \url{https://doi.org/10.18653/v1/D19-3002}.

\bibitem[Wang et~al.(2020)Wang, Tuyls, Wallace, and
  Singh]{DBLP:conf/emnlp/WangTW020}
Junlin Wang, Jens Tuyls, Eric Wallace, and Sameer Singh.
\newblock Gradient-based analysis of {NLP} models is manipulable.
\newblock In Trevor Cohn, Yulan He, and Yang Liu, editors, \emph{Proceedings of
  the 2020 Conference on Empirical Methods in Natural Language Processing:
  Findings, {EMNLP} 2020, Online Event, 16-20 November 2020}, pages 247--258.
  Association for Computational Linguistics, 2020.
\newblock ISBN 978-1-952148-90-3.
\newblock \doi{10.18653/v1/2020.findings-emnlp.24}.
\newblock URL \url{https://doi.org/10.18653/v1/2020.findings-emnlp.24}.

\bibitem[Agarwal et~al.(2020)Agarwal, Frosst, Zhang, Caruana, and
  Hinton]{DBLP:journals/corr/abs-2004-13912}
Rishabh Agarwal, Nicholas Frosst, Xuezhou Zhang, Rich Caruana, and Geoffrey~E.
  Hinton.
\newblock Neural additive models: Interpretable machine learning with neural
  nets.
\newblock \emph{CoRR}, abs/2004.13912, 2020.
\newblock URL \url{https://arxiv.org/abs/2004.13912}.

\bibitem[Fankhauser et~al.(2014)Fankhauser, Knappen, and
  Teich]{DBLP:conf/lrec/FankhauserKT14}
Peter Fankhauser, J{\"{o}}rg Knappen, and Elke Teich.
\newblock Exploring and visualizing variation in language resources.
\newblock In Nicoletta Calzolari, Khalid Choukri, Thierry Declerck, Hrafn
  Loftsson, Bente Maegaard, Joseph Mariani, Asunci{\'{o}}n Moreno, Jan Odijk,
  and Stelios Piperidis, editors, \emph{Proceedings of the Ninth International
  Conference on Language Resources and Evaluation, {LREC} 2014, Reykjavik,
  Iceland, May 26-31, 2014}, pages 4125--4128. European Language Resources
  Association {(ELRA)}, 2014.
\newblock URL
  \url{http://www.lrec-conf.org/proceedings/lrec2014/summaries/185.html}.

\bibitem[Kessler(2017)]{DBLP:conf/acl/Kessler17}
Jason~S. Kessler.
\newblock Scattertext: a browser-based tool for visualizing how corpora differ.
\newblock In Mohit Bansal and Heng Ji, editors, \emph{Proceedings of the 55th
  Annual Meeting of the Association for Computational Linguistics, {ACL} 2017,
  Vancouver, Canada, July 30 - August 4, System Demonstrations}, pages 85--90.
  Association for Computational Linguistics, 2017.
\newblock \doi{10.18653/v1/P17-4015}.
\newblock URL \url{https://doi.org/10.18653/v1/P17-4015}.

\bibitem[Karpathy et~al.(2015)Karpathy, Johnson, and
  Li]{DBLP:journals/corr/KarpathyJL15}
Andrej Karpathy, Justin Johnson, and Fei{-}Fei Li.
\newblock Visualizing and understanding recurrent networks.
\newblock \emph{CoRR}, abs/1506.02078, 2015.
\newblock URL \url{http://arxiv.org/abs/1506.02078}.

\bibitem[Luo et~al.(2019)Luo, Jiang, Belinkov, and
  Glass]{DBLP:conf/acl/LuoJBG19}
Hongyin Luo, Lan Jiang, Yonatan Belinkov, and Jim Glass.
\newblock Improving neural language models by segmenting, attending, and
  predicting the future.
\newblock In Anna Korhonen, David~R. Traum, and Lluis Marquez, editors,
  \emph{Proceedings of the 57th Conference of the Association for Computational
  Linguistics, {ACL} 2019, Florence, Italy, July 28- August 2, 2019, Volume 1:
  Long Papers}, pages 1483--1493. Association for Computational Linguistics,
  2019.
\newblock \doi{10.18653/v1/p19-1144}.
\newblock URL \url{https://doi.org/10.18653/v1/p19-1144}.

\bibitem[Lakretz et~al.(2019)Lakretz, Kruszewski, Desbordes, Hupkes, Dehaene,
  and Baroni]{DBLP:conf/naacl/LakretzKDHDB19}
Yair Lakretz, Germ{\'{a}}n Kruszewski, Theo Desbordes, Dieuwke Hupkes,
  Stanislas Dehaene, and Marco Baroni.
\newblock The emergence of number and syntax units in {LSTM} language models.
\newblock In Jill Burstein, Christy Doran, and Thamar Solorio, editors,
  \emph{Proceedings of the 2019 Conference of the North American Chapter of the
  Association for Computational Linguistics: Human Language Technologies,
  {NAACL-HLT} 2019, Minneapolis, MN, USA, June 2-7, 2019, Volume 1 (Long and
  Short Papers)}, pages 11--20. Association for Computational Linguistics,
  2019.
\newblock \doi{10.18653/v1/n19-1002}.
\newblock URL \url{https://doi.org/10.18653/v1/n19-1002}.

\bibitem[Li et~al.(2016)Li, Chen, Hovy, and Jurafsky]{DBLP:conf/naacl/LiCHJ16}
Jiwei Li, Xinlei Chen, Eduard~H. Hovy, and Dan Jurafsky.
\newblock Visualizing and understanding neural models in {NLP}.
\newblock In Kevin Knight, Ani Nenkova, and Owen Rambow, editors, \emph{{NAACL}
  {HLT} 2016, The 2016 Conference of the North American Chapter of the
  Association for Computational Linguistics: Human Language Technologies, San
  Diego California, USA, June 12-17, 2016}, pages 681--691. The Association for
  Computational Linguistics, 2016.
\newblock \doi{10.18653/v1/n16-1082}.
\newblock URL \url{https://doi.org/10.18653/v1/n16-1082}.

\bibitem[Sawatzky et~al.(2019)Sawatzky, Bergner, and
  Popowich]{DBLP:conf/visualization/SawatzkyBP19}
Lindsey Sawatzky, Steven Bergner, and Fred Popowich.
\newblock Visualizing {RNN} states with predictive semantic encodings.
\newblock In \emph{30th {IEEE} Visualization Conference, {IEEE} {VIS} 2019 -
  Short Papers, Vancouver, BC, Canada, October 20-25, 2019}, pages 156--160.
  {IEEE}, 2019.
\newblock ISBN 978-1-7281-4941-7.
\newblock \doi{10.1109/VISUAL.2019.8933744}.
\newblock URL \url{https://doi.org/10.1109/VISUAL.2019.8933744}.

\bibitem[Strobelt et~al.(2018)Strobelt, Gehrmann, Pfister, and
  Rush]{DBLP:journals/tvcg/StrobeltGPR18}
Hendrik Strobelt, Sebastian Gehrmann, Hanspeter Pfister, and Alexander~M. Rush.
\newblock Lstmvis: {A} tool for visual analysis of hidden state dynamics in
  recurrent neural networks.
\newblock \emph{{IEEE} Trans. Vis. Comput. Graph.}, 24\penalty0 (1):\penalty0
  667--676, 2018.
\newblock \doi{10.1109/TVCG.2017.2744158}.
\newblock URL \url{https://doi.org/10.1109/TVCG.2017.2744158}.

\bibitem[Ming et~al.(2017)Ming, Cao, Zhang, Li, Chen, Song, and
  Qu]{DBLP:conf/ieeevast/MingCZLCSQ17}
Yao Ming, Shaozu Cao, Ruixiang Zhang, Zhen Li, Yuanzhe Chen, Yangqiu Song, and
  Huamin Qu.
\newblock Understanding hidden memories of recurrent neural networks.
\newblock In Brian~D. Fisher, Shixia Liu, and Tobias Schreck, editors,
  \emph{12th {IEEE} Conference on Visual Analytics Science and Technology,
  {IEEE} {VAST} 2017, Phoenix, AZ, USA, October 3-6, 2017}, pages 13--24.
  {IEEE} Computer Society, 2017.
\newblock \doi{10.1109/VAST.2017.8585721}.
\newblock URL \url{https://doi.org/10.1109/VAST.2017.8585721}.

\bibitem[Kahng et~al.(2018)Kahng, Andrews, Kalro, and
  Chau]{DBLP:journals/tvcg/KahngAKC18}
Minsuk Kahng, Pierre~Y. Andrews, Aditya Kalro, and Duen Horng~(Polo) Chau.
\newblock Activis: Visual exploration of industry-scale deep neural network
  models.
\newblock \emph{{IEEE} Trans. Vis. Comput. Graph.}, 24\penalty0 (1):\penalty0
  88--97, 2018.
\newblock \doi{10.1109/TVCG.2017.2744718}.
\newblock URL \url{https://doi.org/10.1109/TVCG.2017.2744718}.

\bibitem[Zhang et~al.(2020)Zhang, Yu, Cui, Wu, Wen, and
  Wang]{DBLP:conf/acl/ZhangYCWWW20}
Yufeng Zhang, Xueli Yu, Zeyu Cui, Shu Wu, Zhongzhen Wen, and Liang Wang.
\newblock Every document owns its structure: Inductive text classification via
  graph neural networks.
\newblock In Dan Jurafsky, Joyce Chai, Natalie Schluter, and Joel~R. Tetreault,
  editors, \emph{Proceedings of the 58th Annual Meeting of the Association for
  Computational Linguistics, {ACL} 2020, Online, July 5-10, 2020}, pages
  334--339. Association for Computational Linguistics, 2020.
\newblock \doi{10.18653/v1/2020.acl-main.31}.
\newblock URL \url{https://doi.org/10.18653/v1/2020.acl-main.31}.

\bibitem[Wu et~al.(2020)Wu, Pan, Zhou, Chang, and Zhu]{DBLP:conf/www/WuP0CZ20}
Man Wu, Shirui Pan, Chuan Zhou, Xiaojun Chang, and Xingquan Zhu.
\newblock Unsupervised domain adaptive graph convolutional networks.
\newblock In Yennun Huang, Irwin King, Tie{-}Yan Liu, and Maarten van Steen,
  editors, \emph{{WWW} '20: The Web Conference 2020, Taipei, Taiwan, April
  20-24, 2020}, pages 1457--1467. {ACM} / {IW3C2}, 2020.
\newblock \doi{10.1145/3366423.3380219}.
\newblock URL \url{https://doi.org/10.1145/3366423.3380219}.

\bibitem[Wang and Leskovec(2020)]{DBLP:journals/corr/abs-2002-06755}
Hongwei Wang and Jure Leskovec.
\newblock Unifying graph convolutional neural networks and label propagation.
\newblock \emph{CoRR}, abs/2002.06755, 2020.
\newblock URL \url{https://arxiv.org/abs/2002.06755}.

\bibitem[Fey and Lenssen(2019)]{DBLP:journals/corr/abs-1903-02428}
Matthias Fey and Jan~Eric Lenssen.
\newblock Fast graph representation learning with pytorch geometric.
\newblock \emph{CoRR}, abs/1903.02428, 2019.
\newblock URL \url{http://arxiv.org/abs/1903.02428}.

\bibitem[Kohonen(1995)]{DBLP:books/sp/Kohonen95}
Teuvo Kohonen.
\newblock \emph{Self-Organizing Maps}, volume~30 of \emph{Springer Series in
  Information Sciences}.
\newblock Springer, 1995.
\newblock ISBN 978-3-642-97612-4.
\newblock \doi{10.1007/978-3-642-97610-0}.
\newblock URL \url{https://doi.org/10.1007/978-3-642-97610-0}.

\bibitem[Mandelbrot(1977)]{DBLP:books/fm/Mandelbrot77}
Benoit Mandelbrot.
\newblock \emph{Fractal Geometry of Nature}.
\newblock W. H. Freeman, 1977.

\bibitem[Jain and Wallace(2019)]{DBLP:conf/naacl/JainW19}
Sarthak Jain and Byron~C. Wallace.
\newblock Attention is not explanation.
\newblock In Jill Burstein, Christy Doran, and Thamar Solorio, editors,
  \emph{Proceedings of the 2019 Conference of the North American Chapter of the
  Association for Computational Linguistics: Human Language Technologies,
  {NAACL-HLT} 2019, Minneapolis, MN, USA, June 2-7, 2019, Volume 1 (Long and
  Short Papers)}, pages 3543--3556. Association for Computational Linguistics,
  2019.
\newblock \doi{10.18653/v1/n19-1357}.
\newblock URL \url{https://doi.org/10.18653/v1/n19-1357}.

\bibitem[Wiegreffe and Pinter(2019)]{DBLP:conf/emnlp/WiegreffeP19}
Sarah Wiegreffe and Yuval Pinter.
\newblock Attention is not not explanation.
\newblock In Kentaro Inui, Jing Jiang, Vincent Ng, and Xiaojun Wan, editors,
  \emph{Proceedings of the 2019 Conference on Empirical Methods in Natural
  Language Processing and the 9th International Joint Conference on Natural
  Language Processing, {EMNLP-IJCNLP} 2019, Hong Kong, China, November 3-7,
  2019}, pages 11--20. Association for Computational Linguistics, 2019.
\newblock \doi{10.18653/v1/D19-1002}.
\newblock URL \url{https://doi.org/10.18653/v1/D19-1002}.

\bibitem[Abnar and Zuidema(2020)]{DBLP:conf/acl/AbnarZ20}
Samira Abnar and Willem~H. Zuidema.
\newblock Quantifying attention flow in transformers.
\newblock In Dan Jurafsky, Joyce Chai, Natalie Schluter, and Joel~R. Tetreault,
  editors, \emph{Proceedings of the 58th Annual Meeting of the Association for
  Computational Linguistics, {ACL} 2020, Online, July 5-10, 2020}, pages
  4190--4197. Association for Computational Linguistics, 2020.
\newblock ISBN 978-1-952148-25-5.
\newblock URL \url{https://www.aclweb.org/anthology/2020.acl-main.385/}.

\bibitem[DeRose et~al.(2021)DeRose, Wang, and
  Berger]{DBLP:journals/tvcg/DeRoseWB21}
Joseph~F. DeRose, Jiayao Wang, and Matthew Berger.
\newblock Attention flows: Analyzing and comparing attention mechanisms in
  language models.
\newblock \emph{{IEEE} Trans. Vis. Comput. Graph.}, 27\penalty0 (2):\penalty0
  1160--1170, 2021.
\newblock \doi{10.1109/TVCG.2020.3028976}.
\newblock URL \url{https://doi.org/10.1109/TVCG.2020.3028976}.

\bibitem[Voita et~al.(2019{\natexlab{a}})Voita, Talbot, Moiseev, Sennrich, and
  Titov]{DBLP:conf/acl/VoitaTMST19}
Elena Voita, David Talbot, Fedor Moiseev, Rico Sennrich, and Ivan Titov.
\newblock Analyzing multi-head self-attention: Specialized heads do the heavy
  lifting, the rest can be pruned.
\newblock In Anna Korhonen, David~R. Traum, and Lluis Marquez, editors,
  \emph{Proceedings of the 57th Conference of the Association for Computational
  Linguistics, {ACL} 2019, Florence, Italy, July 28- August 2, 2019, Volume 1:
  Long Papers}, pages 5797--5808. Association for Computational Linguistics,
  2019{\natexlab{a}}.
\newblock \doi{10.18653/v1/p19-1580}.
\newblock URL \url{https://doi.org/10.18653/v1/p19-1580}.

\bibitem[Song et~al.(2020)Song, Wang, Liang, Liu, and
  Jiang]{DBLP:journals/corr/abs-2002-04815}
Youwei Song, Jiahai Wang, Zhiwei Liang, Zhiyue Liu, and Tao Jiang.
\newblock Utilizing {BERT} intermediate layers for aspect based sentiment
  analysis and natural language inference.
\newblock \emph{CoRR}, abs/2002.04815, 2020.
\newblock URL \url{https://arxiv.org/abs/2002.04815}.

\bibitem[Vig et~al.(2020{\natexlab{a}})Vig, Gehrmann, Belinkov, Qian, Nevo,
  Singer, and Shieber]{DBLP:journals/corr/abs-2004-12265}
Jesse Vig, Sebastian Gehrmann, Yonatan Belinkov, Sharon Qian, Daniel Nevo,
  Yaron Singer, and Stuart~M. Shieber.
\newblock Causal mediation analysis for interpreting neural {NLP:} the case of
  gender bias.
\newblock \emph{CoRR}, abs/2004.12265, 2020{\natexlab{a}}.
\newblock URL \url{https://arxiv.org/abs/2004.12265}.

\bibitem[Voita et~al.(2019{\natexlab{b}})Voita, Sennrich, and
  Titov]{DBLP:conf/emnlp/VoitaST19a}
Elena Voita, Rico Sennrich, and Ivan Titov.
\newblock The bottom-up evolution of representations in the transformer: {A}
  study with machine translation and language modeling objectives.
\newblock In Kentaro Inui, Jing Jiang, Vincent Ng, and Xiaojun Wan, editors,
  \emph{Proceedings of the 2019 Conference on Empirical Methods in Natural
  Language Processing and the 9th International Joint Conference on Natural
  Language Processing, {EMNLP-IJCNLP} 2019, Hong Kong, China, November 3-7,
  2019}, pages 4395--4405. Association for Computational Linguistics,
  2019{\natexlab{b}}.
\newblock \doi{10.18653/v1/D19-1448}.
\newblock URL \url{https://doi.org/10.18653/v1/D19-1448}.

\bibitem[Tenney et~al.(2019)Tenney, Das, and Pavlick]{DBLP:conf/acl/TenneyDP19}
Ian Tenney, Dipanjan Das, and Ellie Pavlick.
\newblock {BERT} rediscovers the classical {NLP} pipeline.
\newblock In Anna Korhonen, David~R. Traum, and Lluis Marquez, editors,
  \emph{Proceedings of the 57th Conference of the Association for Computational
  Linguistics, {ACL} 2019, Florence, Italy, July 28- August 2, 2019, Volume 1:
  Long Papers}, pages 4593--4601. Association for Computational Linguistics,
  2019.
\newblock \doi{10.18653/v1/p19-1452}.
\newblock URL \url{https://doi.org/10.18653/v1/p19-1452}.

\bibitem[Dufter and Sch{\"{u}}tze(2020)]{DBLP:journals/corr/abs-2005-00396}
Philipp Dufter and Hinrich Sch{\"{u}}tze.
\newblock Identifying necessary elements for bert's multilinguality.
\newblock \emph{CoRR}, abs/2005.00396, 2020.
\newblock URL \url{https://arxiv.org/abs/2005.00396}.

\bibitem[Eger et~al.(2020)Eger, Daxenberger, and
  Gurevych]{DBLP:conf/conll/EgerDG20}
Steffen Eger, Johannes Daxenberger, and Iryna Gurevych.
\newblock How to probe sentence embeddings in low-resource languages: On
  structural design choices for probing task evaluation.
\newblock In Raquel Fern{\'{a}}ndez and Tal Linzen, editors, \emph{Proceedings
  of the 24th Conference on Computational Natural Language Learning, CoNLL
  2020, Online, November 19-20, 2020}, pages 108--118. Association for
  Computational Linguistics, 2020.
\newblock \doi{10.18653/v1/2020.conll-1.8}.
\newblock URL \url{https://doi.org/10.18653/v1/2020.conll-1.8}.

\bibitem[Voita and
  Titov(2020{\natexlab{a}})]{DBLP:journals/corr/abs-2003-12298}
Elena Voita and Ivan Titov.
\newblock Information-theoretic probing with minimum description length.
\newblock \emph{CoRR}, abs/2003.12298, 2020{\natexlab{a}}.
\newblock URL \url{https://arxiv.org/abs/2003.12298}.

\bibitem[Gauthier et~al.(2020)Gauthier, Hu, Wilcox, Qian, and
  Levy]{DBLP:conf/acl/GauthierHWQL20}
Jon Gauthier, Jennifer Hu, Ethan Wilcox, Peng Qian, and Roger Levy.
\newblock Syntaxgym: An online platform for targeted evaluation of language
  models.
\newblock In Asli {\c{C}}elikyilmaz and Tsung{-}Hsien Wen, editors,
  \emph{Proceedings of the 58th Annual Meeting of the Association for
  Computational Linguistics: System Demonstrations, {ACL} 2020, Online, July
  5-10, 2020}, pages 70--76. Association for Computational Linguistics, 2020.
\newblock ISBN 978-1-952148-04-0.
\newblock \doi{10.18653/v1/2020.acl-demos.10}.
\newblock URL \url{https://doi.org/10.18653/v1/2020.acl-demos.10}.

\bibitem[Reif et~al.(2019)Reif, Yuan, Wattenberg, Vi{\'{e}}gas, Coenen, Pearce,
  and Kim]{DBLP:conf/nips/ReifYWVCPK19}
Emily Reif, Ann Yuan, Martin Wattenberg, Fernanda~B. Vi{\'{e}}gas, Andy Coenen,
  Adam Pearce, and Been Kim.
\newblock Visualizing and measuring the geometry of {BERT}.
\newblock In Hanna~M. Wallach, Hugo Larochelle, Alina Beygelzimer, Florence
  d'Alch{\'{e}}{-}Buc, Emily~B. Fox, and Roman Garnett, editors, \emph{Advances
  in Neural Information Processing Systems 32: Annual Conference on Neural
  Information Processing Systems 2019, NeurIPS 2019, December 8-14, 2019,
  Vancouver, BC, Canada}, pages 8592--8600, 2019.
\newblock URL
  \url{https://proceedings.neurips.cc/paper/2019/hash/159c1ffe5b61b41b3c4d8f4c2150f6c4-Abstract.html}.

\bibitem[Su et~al.(2020)Su, Zhu, Cao, Li, Lu, Wei, and
  Dai]{DBLP:conf/iclr/SuZCLLWD20}
Weijie Su, Xizhou Zhu, Yue Cao, Bin Li, Lewei Lu, Furu Wei, and Jifeng Dai.
\newblock {VL-BERT:} pre-training of generic visual-linguistic representations.
\newblock In \emph{8th International Conference on Learning Representations,
  {ICLR} 2020, Addis Ababa, Ethiopia, April 26-30, 2020}. OpenReview.net, 2020.
\newblock URL \url{https://openreview.net/forum?id=SygXPaEYvH}.

\bibitem[Conneau et~al.(2018)Conneau, Kruszewski, Lample, Barrault, and
  Baroni]{DBLP:conf/acl/BaroniBLKC18}
Alexis Conneau, Germ{\'{a}}n Kruszewski, Guillaume Lample, Lo{\"{\i}}c
  Barrault, and Marco Baroni.
\newblock What you can cram into a single {\textbackslash}{\&}!{\#}* vector:
  Probing sentence embeddings for linguistic properties.
\newblock In Iryna Gurevych and Yusuke Miyao, editors, \emph{Proceedings of the
  56th Annual Meeting of the Association for Computational Linguistics, {ACL}
  2018, Melbourne, Australia, July 15-20, 2018, Volume 1: Long Papers}, pages
  2126--2136. Association for Computational Linguistics, 2018.
\newblock ISBN 978-1-948087-32-2.
\newblock \doi{10.18653/v1/P18-1198}.
\newblock URL \url{https://www.aclweb.org/anthology/P18-1198/}.

\bibitem[Hewitt and Manning(2019)]{DBLP:conf/naacl/HewittM19}
John Hewitt and Christopher~D. Manning.
\newblock A structural probe for finding syntax in word representations.
\newblock In Jill Burstein, Christy Doran, and Thamar Solorio, editors,
  \emph{Proceedings of the 2019 Conference of the North American Chapter of the
  Association for Computational Linguistics: Human Language Technologies,
  {NAACL-HLT} 2019, Minneapolis, MN, USA, June 2-7, 2019, Volume 1 (Long and
  Short Papers)}, pages 4129--4138. Association for Computational Linguistics,
  2019.
\newblock \doi{10.18653/v1/n19-1419}.
\newblock URL \url{https://doi.org/10.18653/v1/n19-1419}.

\bibitem[Maudslay et~al.(2020)Maudslay, Valvoda, Pimentel, Williams, and
  Cotterell]{DBLP:journals/corr/abs-2005-01641}
Rowan~Hall Maudslay, Josef Valvoda, Tiago Pimentel, Adina Williams, and Ryan
  Cotterell.
\newblock A tale of a probe and a parser.
\newblock \emph{CoRR}, abs/2005.01641, 2020.
\newblock URL \url{https://arxiv.org/abs/2005.01641}.

\bibitem[Ettinger et~al.(2016)Ettinger, Elgohary, and
  Resnik]{DBLP:conf/repeval/EttingerER16}
Allyson Ettinger, Ahmed Elgohary, and Philip Resnik.
\newblock Probing for semantic evidence of composition by means of simple
  classification tasks.
\newblock In \emph{Proceedings of the 1st Workshop on Evaluating Vector-Space
  Representations for NLP, RepEval@ACL 2016, Berlin, Germany, August 2016},
  pages 134--139. Association for Computational Linguistics, 2016.
\newblock \doi{10.18653/v1/W16-2524}.
\newblock URL \url{https://doi.org/10.18653/v1/W16-2524}.

\bibitem[Voita and Titov(2020{\natexlab{b}})]{DBLP:conf/emnlp/VoitaT20}
Elena Voita and Ivan Titov.
\newblock Information-theoretic probing with minimum description length.
\newblock In Bonnie Webber, Trevor Cohn, Yulan He, and Yang Liu, editors,
  \emph{Proceedings of the 2020 Conference on Empirical Methods in Natural
  Language Processing, {EMNLP} 2020, Online, November 16-20, 2020}, pages
  183--196. Association for Computational Linguistics, 2020{\natexlab{b}}.
\newblock \doi{10.18653/v1/2020.emnlp-main.14}.
\newblock URL \url{https://doi.org/10.18653/v1/2020.emnlp-main.14}.

\bibitem[Pilault et~al.(2020)Pilault, Park, and
  Pal]{DBLP:journals/corr/abs-2002-09084}
Jonathan Pilault, Jaehong Park, and Christopher~J. Pal.
\newblock On the impressive performance of randomly weighted encoders in
  summarization tasks.
\newblock \emph{CoRR}, abs/2002.09084, 2020.
\newblock URL \url{https://arxiv.org/abs/2002.09084}.

\bibitem[Vig(2019{\natexlab{b}})]{DBLP:journals/corr/abs-1904-02679}
Jesse Vig.
\newblock Visualizing attention in transformer-based language representation
  models.
\newblock \emph{CoRR}, abs/1904.02679, 2019{\natexlab{b}}.
\newblock URL \url{http://arxiv.org/abs/1904.02679}.

\bibitem[van Aken et~al.(2020)van Aken, Winter, L{\"{o}}ser, and
  Gers]{DBLP:conf/www/AkenWLG20}
Betty van Aken, Benjamin Winter, Alexander L{\"{o}}ser, and Felix~A. Gers.
\newblock Visbert: Hidden-state visualizations for transformers.
\newblock In Amal El~Fallah Seghrouchni, Gita Sukthankar, Tie{-}Yan Liu, and
  Maarten van Steen, editors, \emph{Companion of The 2020 Web Conference 2020,
  Taipei, Taiwan, April 20-24, 2020}, pages 207--211. {ACM} / {IW3C2}, 2020.
\newblock ISBN 978-1-4503-7024-0.
\newblock \doi{10.1145/3366424.3383542}.
\newblock URL \url{https://doi.org/10.1145/3366424.3383542}.

\bibitem[Skrlj et~al.(2020)Skrlj, Erzen, Sheehan, Luz, Robnik{-}Sikonja, and
  Pollak]{DBLP:journals/corr/abs-2005-05716}
Blaz Skrlj, Nika Erzen, Shane Sheehan, Saturnino Luz, Marko Robnik{-}Sikonja,
  and Senja Pollak.
\newblock Attviz: Online exploration of self-attention for transparent neural
  language modeling.
\newblock \emph{CoRR}, abs/2005.05716, 2020.
\newblock URL \url{https://arxiv.org/abs/2005.05716}.

\bibitem[Kobayashi et~al.(2020)Kobayashi, Kuribayashi, Yokoi, and
  Inui]{DBLP:journals/corr/abs-2004-10102}
Goro Kobayashi, Tatsuki Kuribayashi, Sho Yokoi, and Kentaro Inui.
\newblock Attention module is not only a weight: Analyzing transformers with
  vector norms.
\newblock \emph{CoRR}, abs/2004.10102, 2020.
\newblock URL \url{https://arxiv.org/abs/2004.10102}.

\bibitem[Yun et~al.(2021)Yun, Chen, Olshausen, and
  LeCun]{DBLP:journals/corr/abs-2103-15949}
Zeyu Yun, Yubei Chen, Bruno~A. Olshausen, and Yann LeCun.
\newblock Transformer visualization via dictionary learning: contextualized
  embedding as a linear superposition of transformer factors.
\newblock \emph{CoRR}, abs/2103.15949, 2021.
\newblock URL \url{https://arxiv.org/abs/2103.15949}.

\bibitem[Brasoveanu et~al.(2018)Brasoveanu, Rizzo, Kuntschik, Weichselbraun,
  and Nixon]{DBLP:conf/lrec/00020KWN18}
Adrian Brasoveanu, Giuseppe Rizzo, Philipp Kuntschik, Albert Weichselbraun, and
  Lyndon J.~B. Nixon.
\newblock Framing named entity linking error types.
\newblock In Nicoletta Calzolari, Khalid Choukri, Christopher Cieri, Thierry
  Declerck, Sara Goggi, K{\^{o}}iti Hasida, Hitoshi Isahara, Bente Maegaard,
  Joseph Mariani, H{\'{e}}l{\`{e}}ne Mazo, Asunci{\'{o}}n Moreno, Jan Odijk,
  Stelios Piperidis, and Takenobu Tokunaga, editors, \emph{Proceedings of the
  Eleventh International Conference on Language Resources and Evaluation,
  {LREC} 2018, Miyazaki, Japan, May 7-12, 2018}. European Language Resources
  Association {(ELRA)}, 2018.
\newblock URL
  \url{http://www.lrec-conf.org/proceedings/lrec2018/summaries/612.html}.

\bibitem[van~der Heijden et~al.(2020)van~der Heijden, Abnar, and
  Shutova]{DBLP:conf/aaai/HeijdenAS20}
Niels van~der Heijden, Samira Abnar, and Ekaterina Shutova.
\newblock A comparison of architectures and pretraining methods for
  contextualized multilingual word embeddings.
\newblock In \emph{The Thirty-Fourth {AAAI} Conference on Artificial
  Intelligence, {AAAI} 2020, The Thirty-Second Innovative Applications of
  Artificial Intelligence Conference, {IAAI} 2020, The Tenth {AAAI} Symposium
  on Educational Advances in Artificial Intelligence, {EAAI} 2020, New York,
  NY, USA, February 7-12, 2020}, pages 9090--9097. {AAAI} Press, 2020.
\newblock URL \url{https://aaai.org/ojs/index.php/AAAI/article/view/6443}.

\bibitem[Gan et~al.(2020)Gan, Chen, Li, Zhu, Cheng, and
  Liu]{DBLP:conf/nips/Gan0LZ0020}
Zhe Gan, Yen{-}Chun Chen, Linjie Li, Chen Zhu, Yu~Cheng, and Jingjing Liu.
\newblock Large-scale adversarial training for vision-and-language
  representation learning.
\newblock In Hugo Larochelle, Marc'Aurelio Ranzato, Raia Hadsell,
  Maria{-}Florina Balcan, and Hsuan{-}Tien Lin, editors, \emph{Advances in
  Neural Information Processing Systems 33: Annual Conference on Neural
  Information Processing Systems 2020, NeurIPS 2020, December 6-12, 2020,
  virtual}, 2020.
\newblock URL
  \url{https://proceedings.neurips.cc/paper/2020/hash/49562478de4c54fafd4ec46fdb297de5-Abstract.html}.

\bibitem[Cao et~al.(2020)Cao, Gan, Cheng, Yu, Chen, and
  Liu]{DBLP:conf/eccv/CaoGCY0020}
Jize Cao, Zhe Gan, Yu~Cheng, Licheng Yu, Yen{-}Chun Chen, and Jingjing Liu.
\newblock Behind the scene: Revealing the secrets of pre-trained
  vision-and-language models.
\newblock In Andrea Vedaldi, Horst Bischof, Thomas Brox, and Jan{-}Michael
  Frahm, editors, \emph{Computer Vision - {ECCV} 2020 - 16th European
  Conference, Glasgow, UK, August 23-28, 2020, Proceedings, Part {VI}}, volume
  12351 of \emph{Lecture Notes in Computer Science}, pages 565--580. Springer,
  2020.
\newblock ISBN 978-3-030-58538-9.
\newblock \doi{10.1007/978-3-030-58539-6\_34}.
\newblock URL \url{https://doi.org/10.1007/978-3-030-58539-6\_34}.

\bibitem[Han et~al.(2021)Han, Xiao, Wu, Guo, Xu, and
  Wang]{DBLP:journals/corr/abs-2103-00112}
Kai Han, An~Xiao, Enhua Wu, Jianyuan Guo, Chunjing Xu, and Yunhe Wang.
\newblock Transformer in transformer.
\newblock \emph{CoRR}, abs/2103. 00112, 2021.
\newblock URL \url{https://arxiv.org/abs/2103.00112}.

\bibitem[Li et~al.(2020)Li, Wang, and Luo]{DBLP:conf/bibm/LiW020}
Yikuan Li, Hanyin Wang, and Yuan Luo.
\newblock A comparison of pre-trained vision-and-language models for multimodal
  representation learning across medical images and reports.
\newblock In Taesung Park, Young{-}Rae Cho, Xiaohua Hu, Illhoi Yoo, Hyun~Goo
  Woo, Jianxin Wang, Julio~C. Facelli, Seungyoon Nam, and Mingon Kang, editors,
  \emph{{IEEE} International Conference on Bioinformatics and Biomedicine,
  {BIBM} 2020, Virtual Event, South Korea, December 16-19, 2020}, pages
  1999--2004. {IEEE}, 2020.
\newblock ISBN 978-1-7281-6215-7.
\newblock \doi{10.1109/BIBM49941.2020.9313289}.
\newblock URL \url{https://doi.org/10.1109/BIBM49941.2020.9313289}.

\bibitem[Kim et~al.(2021)Kim, Son, and Kim]{DBLP:conf/icml/KimSK21}
Wonjae Kim, Bokyung Son, and Ildoo Kim.
\newblock Vilt: Vision - and - language transformer without convolution or
  region supervision.
\newblock In Marina Meila and Tong Zhang, editors, \emph{Proceedings of the
  38th International Conference on Machine Learning, {ICML} 2021, 18-24 July
  2021, Virtual Event}, volume 139 of \emph{Proceedings of Machine Learning
  Research}, pages 5583--5594. {PMLR}, 2021.
\newblock URL \url{http://proceedings.mlr.press/v139/kim21k.html}.

\bibitem[Zhong et~al.(2019)Zhong, Wang, Liu, Qiu, and
  Huang]{DBLP:journals/corr/abs-1909-13705}
Ming Zhong, Danqing Wang, Pengfei Liu, Xipeng Qiu, and Xuanjing Huang.
\newblock A closer look at data bias in neural extractive summarization models.
\newblock \emph{CoRR}, abs/1909.13705, 2019.
\newblock URL \url{http://arxiv.org/abs/1909.13705}.

\bibitem[Slack et~al.(2019)Slack, Hilgard, Jia, Singh, and
  Lakkaraju]{DBLP:journals/corr/abs-1911-02508}
Dylan Slack, Sophie Hilgard, Emily Jia, Sameer Singh, and Himabindu Lakkaraju.
\newblock How can we fool {LIME} and shap? adversarial attacks on post hoc
  explanation methods.
\newblock \emph{CoRR}, abs/1911.02508, 2019.
\newblock URL \url{http://arxiv.org/abs/1911.02508}.

\bibitem[Han et~al.(2019)Han, Gao, Yao, Ye, Liu, and
  Sun]{DBLP:conf/emnlp/HanGYYLS19}
Xu~Han, Tianyu Gao, Yuan Yao, Deming Ye, Zhiyuan Liu, and Maosong Sun.
\newblock Opennre: An open and extensible toolkit for neural relation
  extraction.
\newblock In Sebastian Pad{\'{o}} and Ruihong Huang, editors, \emph{Proceedings
  of the 2019 Conference on Empirical Methods in Natural Language Processing
  and the 9th International Joint Conference on Natural Language Processing,
  {EMNLP-IJCNLP} 2019, Hong Kong, China, November 3-7, 2019 - System
  Demonstrations}, pages 169--174. Association for Computational Linguistics,
  2019.
\newblock \doi{10.18653/v1/D19-3029}.
\newblock URL \url{https://doi.org/10.18653/v1/D19-3029}.

\bibitem[Chen et~al.(2019)Chen, Khashabi, Yin, Callison{-}Burch, and
  Roth]{DBLP:conf/naacl/ChenK0CR19}
Sihao Chen, Daniel Khashabi, Wenpeng Yin, Chris Callison{-}Burch, and Dan Roth.
\newblock Seeing things from a different angle: Discovering diverse
  perspectives about claims.
\newblock In Jill Burstein, Christy Doran, and Thamar Solorio, editors,
  \emph{Proceedings of the 2019 Conference of the North American Chapter of the
  Association for Computational Linguistics: Human Language Technologies,
  {NAACL-HLT} 2019, Minneapolis, MN, USA, June 2-7, 2019, Volume 1 (Long and
  Short Papers)}, pages 542--557. Association for Computational Linguistics,
  2019.
\newblock \doi{10.18653/v1/n19-1053}.
\newblock URL \url{https://doi.org/10.18653/v1/n19-1053}.

\bibitem[Wadden et~al.(2019)Wadden, Wennberg, Luan, and
  Hajishirzi]{DBLP:conf/emnlp/WaddenWLH19}
David Wadden, Ulme Wennberg, Yi~Luan, and Hannaneh Hajishirzi.
\newblock Entity, relation, and event extraction with contextualized span
  representations.
\newblock In Kentaro Inui, Jing Jiang, Vincent Ng, and Xiaojun Wan, editors,
  \emph{Proceedings of the 2019 Conference on Empirical Methods in Natural
  Language Processing and the 9th International Joint Conference on Natural
  Language Processing, {EMNLP-IJCNLP} 2019, Hong Kong, China, November 3-7,
  2019}, pages 5783--5788. Association for Computational Linguistics, 2019.
\newblock \doi{10.18653/v1/D19-1585}.
\newblock URL \url{https://doi.org/10.18653/v1/D19-1585}.

\bibitem[Gao et~al.(2019)Gao, Han, Zhu, Liu, Li, Sun, and
  Zhou]{DBLP:conf/emnlp/GaoHZLLSZ19}
Tianyu Gao, Xu~Han, Hao Zhu, Zhiyuan Liu, Peng Li, Maosong Sun, and Jie Zhou.
\newblock Fewrel 2.0: Towards more challenging few-shot relation
  classification.
\newblock In Kentaro Inui, Jing Jiang, Vincent Ng, and Xiaojun Wan, editors,
  \emph{Proceedings of the 2019 Conference on Empirical Methods in Natural
  Language Processing and the 9th International Joint Conference on Natural
  Language Processing, {EMNLP-IJCNLP} 2019, Hong Kong, China, November 3-7,
  2019}, pages 6249--6254. Association for Computational Linguistics, 2019.
\newblock \doi{10.18653/v1/D19-1649}.
\newblock URL \url{https://doi.org/10.18653/v1/D19-1649}.

\bibitem[Gonen and Goldberg(2019)]{DBLP:conf/acl-wnlp/GonenG19}
Hila Gonen and Yoav Goldberg.
\newblock Lipstick on a pig: Debiasing methods cover up systematic gender
  biases in word embeddings but do not remove them.
\newblock In Amittai Axelrod, Diyi Yang, Rossana Cunha, Samira Shaikh, and
  Zeerak Waseem, editors, \emph{Proceedings of the 2019 Workshop on Widening
  NLP@ACL 2019, Florence, Italy, July 28, 2019}, pages 60--63. Association for
  Computational Linguistics, 2019.
\newblock ISBN 978-1-950737-42-0.
\newblock URL \url{https://www.aclweb.org/anthology/W19-3621/}.

\bibitem[Vig et~al.(2020{\natexlab{b}})Vig, Madani, Varshney, Xiong, Socher,
  and Rajani]{DBLP:journals/corr/abs-2006-15222}
Jesse Vig, Ali Madani, Lav~R. Varshney, Caiming Xiong, Richard Socher, and
  Nazneen~Fatema Rajani.
\newblock Bertology meets biology: Interpreting attention in protein language
  models.
\newblock \emph{CoRR}, abs/2006.15222, 2020{\natexlab{b}}.
\newblock URL \url{https://arxiv.org/abs/2006.15222}.

\end{thebibliography}





% \end{thebibliography}

\end{document}